%% file: acl_latex.tex
\pdfoutput=1

\documentclass[11pt]{article}

\usepackage[final]{acl}

\usepackage{times}
\usepackage{latexsym}

\usepackage[T1]{fontenc}

\usepackage[utf8]{inputenc}

\usepackage{microtype}

\usepackage{inconsolata}

\usepackage{graphicx}

\usepackage{kotex}
\usepackage{multirow} 
\usepackage{booktabs}
\usepackage{amsmath}
\usepackage{array}
\usepackage{pdfpages}
\usepackage{subcaption}
\usepackage{longtable}
\usepackage{placeins}
\usepackage{afterpage}  
\usepackage{caption} 
\usepackage{multicol}
\usepackage{lineno}  

\newcommand{\ours}{\textsc{CREFT}}

\title{\ours{}: Sequential Multi-Agent LLM for \\ Character Relation Extraction}

\input{sections/00_authors}

\begin{document}
\maketitle
\input{sections/01_abstract}

\input{sections/02_introduction}

\input{sections/03_related_work}

\input{sections/04_framework}

\input{sections/05_experiments}

\input{sections/06_results_and_analysis}

\input{sections/08_conclusion}

\bibliography{custom}

\appendix
\onecolumn
\input{sections/10_appendix}

\end{document}

%% file: sections/00_authors.tex
\author{
    Ye Eun Chun\\
    CJ Corporation\\
    \And
    Taeyoon Hwang\\
    CJ Corporation\\
    \And
    Seung-won Hwang\\
    Seoul National University\\
    \And
    Byung-Hak Kim\thanks{Corresponding author. Correspondence to: \texttt{bhak.kim@cj.net}}\\
    CJ Corporation\\
}

%% file: sections/01_abstract.tex
\begin{abstract}
Understanding complex character relations is crucial for narrative analysis and efficient script evaluation, yet existing extraction methods often fail to handle long-form narratives with nuanced interactions. To address this challenge, we present \textbf{\ours{}}, a novel sequential framework leveraging specialized Large Language Model (LLM) agents. First, \ours{} builds a base character graph through knowledge distillation, then iteratively refines character composition, relation extraction, role identification, and group assignments. Experiments on a curated Korean drama dataset demonstrate that \ours{} significantly outperforms single-agent LLM baselines in both accuracy and completeness. By systematically visualizing character networks, \ours{} streamlines narrative comprehension and accelerates script review—offering substantial benefits to the entertainment, publishing, and educational sectors.
\end{abstract}

%% file: sections/02_introduction.tex
\section{Introduction}
The rapid global growth of entertainment content—dramas, films, and serialized narratives—has led to a massive increase in scripts requiring efficient review. Rapidly grasping complex character relationships in these scripts is crucial for content selection, production planning, and audience engagement~\citep{11_elson-etal-2010-extracting,3_llmfallshort}. However, existing automated approaches often fall short: they struggle with coreference resolution, miss nuanced or implicit interactions, and fail to handle large-scale, long-form narratives~\citep{niraula-etal-2014-dare, liu2023survey}.

To overcome these limitations, we present \textbf{\ours{}}, a \emph{sequential}, \emph{multi-agent} framework leveraging specialized LLM agents. Unlike single-pass methods, \ours{} processes narrative data \emph{iteratively}: each agent refines one aspect—character composition, explicit/implicit relation extraction, role identification, or group assignments—thus systematically enhancing accuracy. Our evaluations on a human-annotated Korean drama dataset demonstrate significant gains, including a 18.8\% absolute improvement in character recall and a 16.5\% boost in group match F1-score compared to single-pass baselines.

Specifically, this paper addresses:
\begin{enumerate}
    \item How sequential, agent-by-agent refinement improves character identity resolution,
    \item To what extent it enhances the detection of explicit and implicit relationships,
    \item Whether it effectively solves persistent challenges of group assignments in ambiguous or evolving narratives.
\end{enumerate}
Our main contributions are:
\begin{enumerate}
    \item A clearly defined Character Relation Structure (CRS) for efficient, visually intuitive narrative analysis,
    \item A novel sequential multi-agent LLM refinement pipeline that significantly boosts extraction accuracy and completeness.
\end{enumerate}
In short, \ours{} supplies a powerful, scalable solution for character-centric narrative analysis, offering notable benefits for script evaluation in entertainment, publishing, and educational contexts while advancing entity and relation extraction in Natural Language Processing.

%% file: sections/03_related_work.tex
\section{Related Work}
\subsection{Traditional Approaches to Character Relation Extraction}
Early methods for character relation extraction heavily relied on manual or heuristic-based techniques~\citep{1_fictionalcharacterNet, 8_nazi}, which offered precision but were highly labor-intensive and thus impractical for large-scale narratives. To address scalability, automated co-occurrence networks~\citep{alberich2002marveluniverselookslike} assumed that characters mentioned together shared relationships. However, these undirected networks often introduced noise (mere co-occurrence does not guarantee interaction) and lacked nuanced directional or emotional attributes. Subsequently, dialogue-based networks emerged to capture explicit exchanges between characters \citep{11_elson-etal-2010-extracting}, although they struggled to detect implicit relationships embedded in descriptive text. Later refinements~\citep{9_lee-yeung-2012-extracting} added proximity-based connections even without direct quotes, partially mitigating the limitations of strictly quotation-driven approaches. Nevertheless, early automated methods often sacrificed detail and nuance, motivating more advanced solutions in subsequent research.

\subsection{Structuring Character Relations}
Beyond extraction, structuring character relationships poses additional challenges. Early work relied on network-based clustering (e.g., using edge betweenness to detect communities \cite{7_communitystructure}), but these methods mainly performed structural partitioning, overlooking the critical issue of character identification. To address identity resolution, \cite{10_chen-etal-2017-robust} employed a neural network model linking name variants and pronouns to the same character, achieving high coreference accuracy. Moreover, characterization often extends beyond explicit interactions. For instance, \cite{4_personas} introduced a latent variable model that groups characters by internal traits, roles, and implicit functions—offering deeper insights into narrative complexity than simple relationship graphs. Together, these studies highlight the importance of robust identity resolution and persona modeling for effective character-relationship structuring.

\subsection{LLM-Based Character Relation Extraction}
Recent advances in LLMs have substantially improved character relation extraction by capturing nuanced textual cues, implicit context, and broad world knowledge. Benchmarks like Conan~\cite{3_llmfallshort} reveal that while models such as GPT-4~\cite{openai2023gpt4} and Llama2~\cite{touvron2023llama} effectively parse explicit relationships, they struggle with implicit or multi-layered dynamics due to limited context windows. Prompt-engineering techniques~\cite{2_attributefrommovie} and problem decomposition (e.g., chain-of-thought prompting) have mitigated some issues, but challenges remain: LLMs may fabricate non-existent ties or miss subtle relations. In response, \ours{} leverages an iterative multi-agent pipeline, wherein specialized LLM agents each refine distinct tasks—addressing context loss, ensuring reliable extraction of implicit relations, and enforcing consistent character grouping—ultimately enhancing accuracy and usability for long-form narratives.

%% file: sections/04_framework.tex
\input{figures/02_creft}
\section{CREFT Framework}
The proposed \ours{} framework consists of two main stages: constructing a base character graph and refining it through a sequential chain of specialized LLM agents. Figure \ref{figure_overview} summarizes our approach. The \ours{} framework is specifically tailored to address industry needs during the script review phase, where understanding character dynamics is crucial for early-stage production decisions.\footnote{Since producers typically have limited access to early narrative content—usually only the initial episodes (episodes 1 to 4)—our methodology explicitly accommodates and leverages this restricted scope to provide robust, actionable insights into character interactions and narrative structure.}

\subsection{Base Character Graph Construction} 
The base character graph represents narrative characters as nodes and their relationships as edges. To create this graph, we segment narratives into smaller text chunks and extract Subject-Predicate-Object (SPO) triplets capturing character interactions. 
This process highlights the interaction frequency of characters appearing in episodes 1 to 4, serving as the foundational data for the subsequent refinement steps. 
Our method adapts the knowledge graph construction described by \cite{jeong2025agentasjudgefactualsummarizationlong}, specifically tailored for sequential narrative analysis.

\paragraph{Triplet Extraction via LLM Knowledge Distillation}
To protect script confidentiality prior to their release, direct usage of LLM APIs for character extraction is avoided. Instead, we introduce a secure method utilizing knowledge-distillation. Specifically, we create an annotated training dataset from 355 Korean drama scripts (aired 1995-2023, episodes 1–4), each divided into 512-character segments. Each segment is analyzed by GPT-4o \cite{openai2024gpt4o} to generate SPO triplets, resulting in a dataset comprising 61,483 chunk-response pairs. This dataset is subsequently utilized to fine-tune a specialized Korean-language LLM, specifically optimized for SPO triplet generation. The fine-tuned LLM then processes narrative scripts in sequential chunks and uses a structured prompt shown in Appendix \ref{appendix:triplet_prompt}, systematically generating SPO triplets from each chunk to form the base character graph.

\subsection{Sequential Refinement with LLM Agents}
\label{ssec:seq_refinement}
Accurately reconstructing the CRS from the base character graph requires careful refinement of critical attributes, including character composition, explicit and implicit relations, character roles, and character groups. To systematically address this, we introduce a sequential multi-agent framework consisting of specialized LLM agents, each dedicated to refining specific components of the CRS. Each agent builds upon the output of the previous agent, iteratively improving both accuracy and completeness.\footnote{For a detailed breakdown of each agent, refer to Appendix \ref{appendix:prompt} for the prompts we use, and to Appendix \ref{appendix:stepbystep}
for sample CRS outputs illustrating each refinement stage.}

\paragraph{Character Selection} 
In narrative analysis, key characters are usually introduced early and frequently appear across initial episodes, though this can vary significantly based on genre and storytelling structure. Some essential characters might emerge or gain prominence only as the narrative unfolds, posing challenges to extraction methods based purely on frequency analysis. To address these challenges, our approach begins by collecting user-provided information about main and sub-characters. Using the base character graph, we compute interaction frequencies between character pairs, forming a weighted character relation graph \(G\). Within \(G\), nodes representing main and sub-characters are initially assigned higher importance scores. We then iteratively apply Personalized PageRank (PPR) \cite{ppr} to identify additional characters strongly connected to these prioritized nodes, selecting those that surpass a predefined threshold. Through multiple iterations, we refine and expand the character set, ensuring comprehensive coverage of narratively significant characters.

\paragraph{Merging Duplicate Nodes} 
Characters in narratives often possess multiple identifiers, such as formal names, nicknames, and professional titles. For example, a character named `Young-min Cha' might also be referred as `Professor Cha' depending on context. Representing each identifier as separately can cause confusion and hinder interpretation. To resolve this issue, we employ a specialized LLM agent that leveraged contextual information from treatments and episode summaries to accurately merge nodes referring to the same character. This step ensures precise consolidation of character identities, significantly enhancing the interpretability and clarity of the resulting CRS.

\paragraph{Relation Extraction} 
Character relations can be categorized into explicit and implicit relations. Explicit relations are clearly defined social connections, such as family ties, professional associations, or friendships, typically established during narrative planning and remaining consistent throughout the narrative. In contrast, implicit relations reflect nuanced, emotionally driven, or conflict-based dynamics that evolve over the course of a narrative. 
To effectively capture these relations, we utilize a specialized LLM agent that integrates explicit context with implicit narrative cues—such as predefined emotions—derived from treatments and episode summaries. 
This combined approach ensures a comprehensive understanding of character interactions, effectively capturing both static and evolving relationship dynamics within the narrative.

\paragraph{Filtering Out Irrelevant Characters} 
To enhance clarity in the final CRS, we remove extraneous characters who have minimal impact on the narrative (e.g., transient figures such as patients or customers). First, an LLM classifies these potential candidates from the list generated after the PPR process. Characters with a discernible last name are typically retained to avoid mistakenly discarding story-relevant individuals. Finally, the LLM verifies each flagged entry’s name and narrative context, ensuring only truly irrelevant characters are excluded and essential ones remain for accurate story representation.

\paragraph{Role Identification}
After excluding non-essential characters, we assign roles to each remaining character based on information from the treatment and episode summaries. These roles typically include professions or hierarchical positions but exclude age-related terms or other incidental descriptors.

\paragraph{Grouping Characters}
To present the CRS succinctly yet informatively, we group characters into clusters based on major affiliations or familial ties, while maintaining more direct, higher-information connections for main and sub-characters. Specifically, we employ an LLM to identify pivotal groups—such as family units or organizational teams—from the final character list (after filtering out irrelevant characters), leveraging episode treatments and summaries for context. The assigned groups are then visually distinguished by distinct colors within the CRS, ensuring a clear and interpretable depiction of character relationships.

%% file: figures/02_creft.tex
\begin{figure*}[ht]
    \centering
    \includegraphics[width=1.0\linewidth]{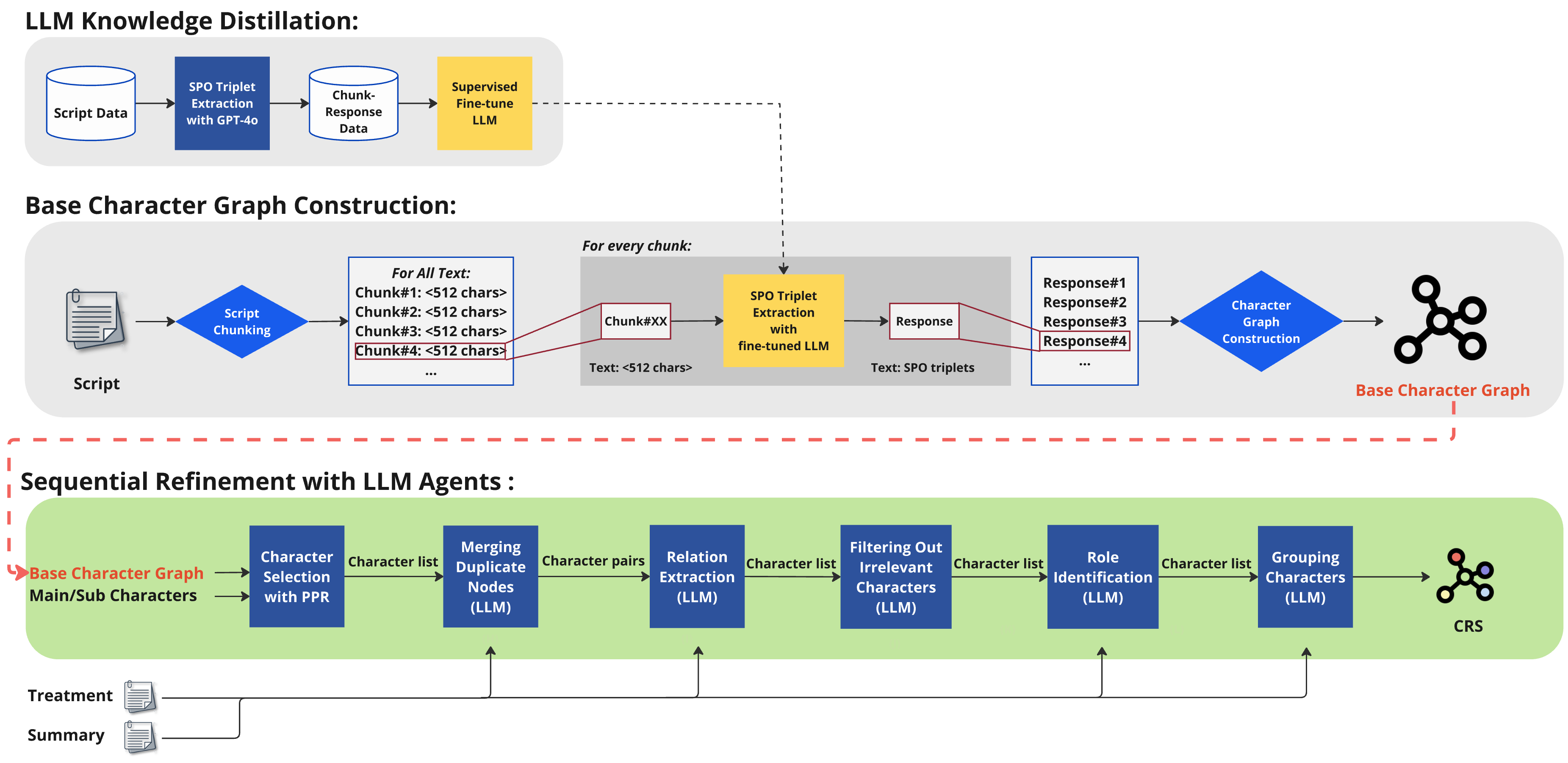}
    \caption{Overview of the \ours{} Framework. The base character graph is generated by extracting SPO triplet data from narrative texts and subsequently refined through specialized LLM agents. Each agent incrementally improves character composition, explicit/implicit relation extraction, role identification, and group assignments, significantly enhancing clarity and accuracy of narrative understanding for early-stage content evaluation.}
    \label{figure_overview}
\end{figure*}

%% file: sections/05_experiments.tex
\section{Experiments}
\subsection{Dataset}
To evaluate the accuracy of our \ours{} framework in generating CRSs, we constructed a custom dataset from our internally developed Korean drama database. We selected 15 previously broadcast dramas (aired from 2014 to 2023) for which complete treatments and scripts were available for episodes 1 to 4. Domain experts in drama analysis then annotated these samples by defining each character’s scope, roles, group affiliations, and key relations between main and sub-characters. 

Because large-scale human annotation is costly and time-consuming, we deliberately chose a modest number of dramas that collectively reflect a diverse set of challenging scenarios. For example, our dataset includes narratives featuring characters who possess others (e.g., dramas 3, 13, 15), multiple characters sharing the same name (e.g., drama 4), and characters who exist across parallel worlds or different timelines (e.g., drama 4, 7, 8, 13). Such complexities introduce additional challenges for character identification and relation extraction beyond standard settings.

\subsection{Evaluation Metrics}
\label{ssec:eval_metrics}
To evaluate our framework’s performance, we adopt several complementary metrics. \textbf{Character recall} measures how comprehensively the correct characters appear in the CRS. For role, group name, explicit relation, and implicit relation, we compute the semantic proximity between predictions and ground truth using cosine similarity with the Korean fine-tuned language model KURE-v1~\cite{KURE}. From these similarity scores, we derive  \textbf{role similarity, group name similarity, explicit relation similarity}, and \textbf{implicit relation similarity}. If multiple roles or relations exist for the same character, we select the pair with the highest similarity score; any role or relation absent in the ground truth is excluded from evaluation. Additionally, \textbf{group match F1-score} assesses the accuracy of character grouping, while \textbf{character-relation recall} measures the correctness of extracted key character pairs. Full details on each metric appear in Appendix~\ref{appendix:metrics}.

\subsection{Experimental Setup}
\paragraph{Experiment 1: Single vs.\ Multi-Agent}
In Experiment~1, we compare the multi-agent approach against a single-agent baseline for generating CRSs, following the PPR-based character selection step. Both approaches rely on the Mistral-Large-Instruct-2411 model by Mistral AI~\cite{mistral2023largeinstruct2411}, which has 123B parameters and supports Korean. In the multi-agent pipeline, each stage (e.g., merging duplicate nodes, relation extraction) is guided by its own prompt, with results passed via regular expressions to the next stage. By contrast, the single-agent approach combines all tasks into a single comprehensive prompt and expects an output containing every required piece of information (e.g., subject/object character names, explicit/implicit relations, roles, groups). We then evaluate both CRSs using the metrics described in Section~\ref{ssec:eval_metrics}. The exact prompt\footnote{The original prompts contain some Korean terms or names (e.g., `교수', meaning 'professor'); their English translations appear in Appendix \ref{appendix:prompt} and \ref{appendix:single_prompt}.} used in Experiment~1 is provided in Appendix~\ref{appendix:prompt} and \ref{appendix:single_prompt}.

\paragraph{Experiment 2: PPR Efficacy}
Experiment~2 evaluates the effectiveness of the PPR-based method for identifying key characters, compared to a simpler approach that ranks characters by node-edge count in the base character graph. In the PPR method, users initially designate main and sub-characters, assigning them importance scores of 1 and 0.5, respectively. We set a threshold of 0.02 to filter out low-relevance characters in the first extraction pass. Subsequently, scores are iteratively updated for each newly identified character (set as 1, and 0 for the rest), again with a 0.02 threshold for determining relevance at each step. The final character list is thus the union of all characters discovered across these iterations. In contrast, the node-edge count method simply ranks characters by the total number of edges each node possesses—then selects the top-scoring nodes equal to the length of the PPR-derived character list. We compare the two methods based on character recall, precision, and F1-score.

%% file: sections/06_results_and_analysis.tex
\input{figures/01_crs}
\input{tables/02_exp1_2}
\input{tables/03_exp2}

\section{Results and Analysis}
\subsection{Quantitative Results}
Table~\ref{tab:comp_summary} summarizes the average scores across seven key metrics for Experiment~1, revealing that the multi-agent approach consistently outperforms the single-agent baseline. Despite this advantage, the multi-agent method exhibits notably low scores (under 10\%) in group match F1-score and group name similarity for at least one drama (Drama~1), and also struggles with implicit relation similarity in another (Drama~12). Meanwhile, the single-agent approach shows broadly poor performance in multiple dramas (Dramas~1, 4, 7, 9, 11, and 14). These patterns are detailed further on a per-drama basis in Appendix~\ref{appendix:experiment_result}.

Among the similarity metrics, \textbf{role similarity} yields the highest average, indicating that role identification is comparatively more straightforward than assigning group affiliations or labeling intricate relationships. Still, the multi-agent framework’s group match F1-score remains below 50\%, suggesting that accurate group assignment remains a persistent challenge. Overall, these findings underscore that a single-agent pipeline struggles with extracting multiple components of a CRS, whereas multi-agent processing delivers more robust results—particularly beyond role identification.

Turning to Experiment~2 (see Table~\ref{tab:ppr_count_summary}), the average precision, recall, and F1 scores for the PPR algorithm closely match those of the node-edge count method, with the latter showing a slight advantage in overall averages. Closer inspection (detailed in Appendix~\ref{appendix:experiment_result2}) reveals that each algorithm wins in five individual dramas, while delivering equivalent results in the others. This suggests that the effectiveness of each method may depend on specific narrative characteristics, such as the presence or frequency of key characters in the initial episodes.

\subsection{Qualitative Results}
Figure~\ref{fig:sidebyside} compares the CRSs generated by single- and multi-agent approaches in Experiment~1. Notably, the single-agent method fails to merge certain characters (e.g., E, O, L and P) that share aliases, resulting in multiple, erroneously distinct nodes (E', O', L', P'). Conversely, the multi-agent approach accurately consolidates these aliases and focuses relationships around the main (largest circle) and sub-characters (second largest circle). Consequently, the multi-agent CRS displays fewer redundant edges, such as extraneous ``colleague'' links.

In Experiment~2, the efficacy of character selection depended on each script's narrative structure. When key characters appeared less frequently in the first four episodes, PPR yielded more accurate results (see Appendix \ref{appendix:ppr_analysis_}). However, in cases where crucial characters were introduced early and appeared frequently, the count-based algorithm proved more effective. This indicates that the optimal extraction strategy may vary based on the drama’s pacing and character introduction patterns.

%% file: figures/01_crs.tex
\begin{figure*}[ht]
    \centering
    \begin{subfigure}[b]{0.49\linewidth}
        \centering
        \includegraphics[width=\linewidth,page=1]{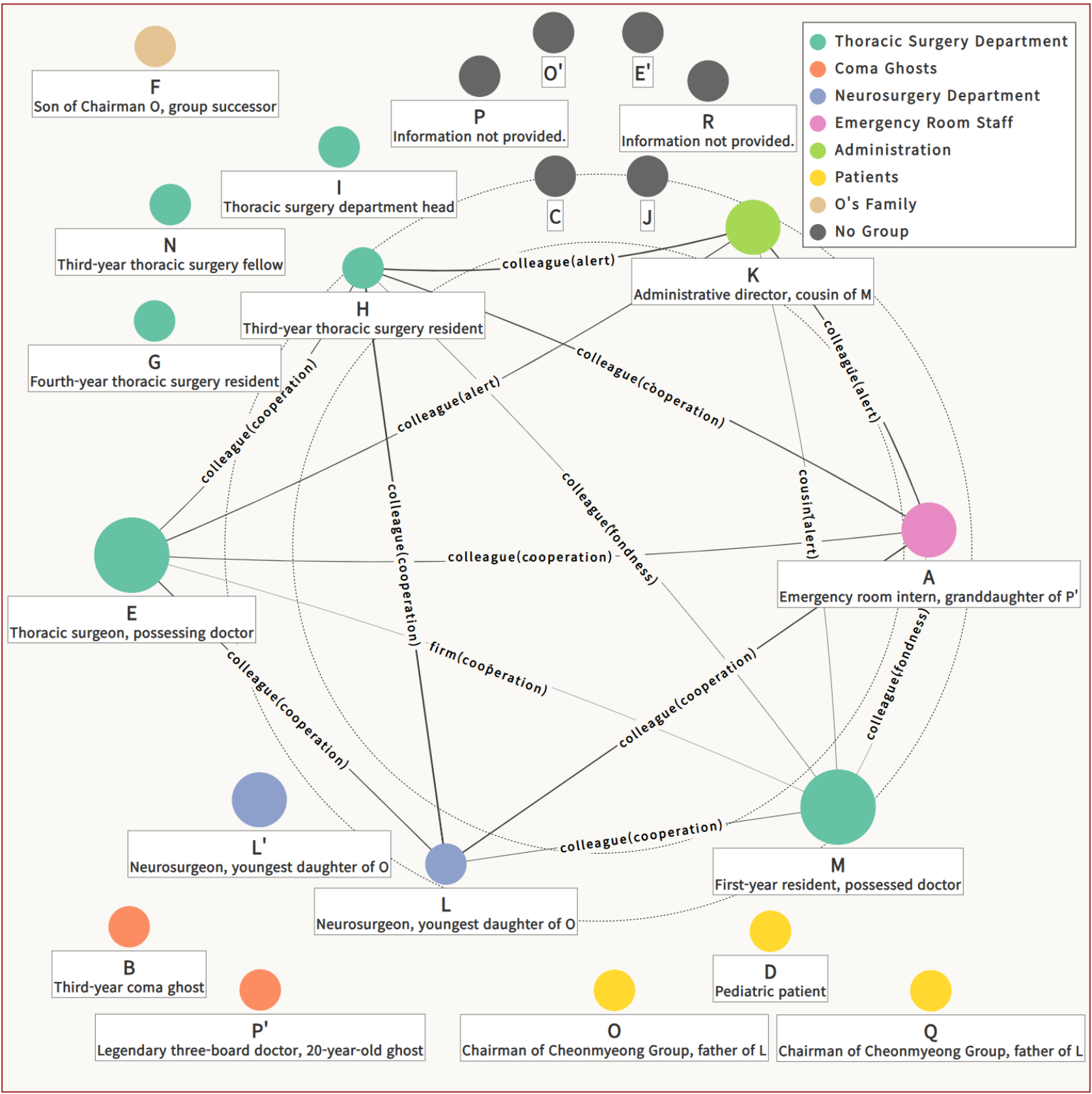}
        \caption{CRS from single-agent approach}
    \end{subfigure}
    \hfill
    \begin{subfigure}[b]{0.49\linewidth}
        \centering
        \includegraphics[width=\linewidth,page=1]{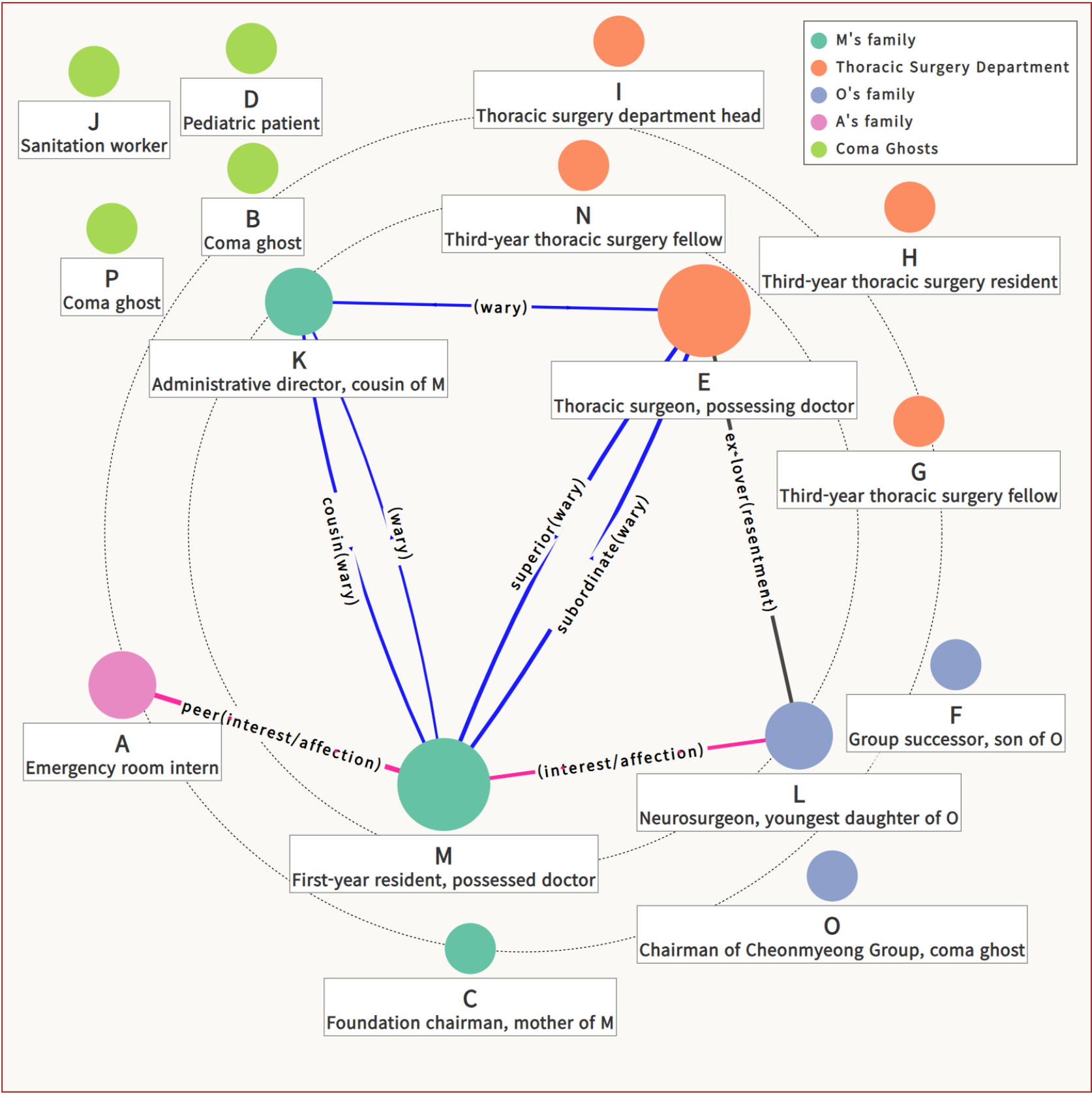}
        \caption{CRS from multi-agent approach}
    \end{subfigure}
    \caption{
    CRSs for \textbf{Drama~13}. In (a), the single-agent approach fails to merge aliases (E, O, L, P), producing redundant nodes (E', O', L', P'). In contrast, (b) shows the multi-agent approach accurately merging these aliases for a more concise representation. 
    }
    \label{fig:sidebyside}
\end{figure*}

%% file: tables/02_exp1_2.tex
\begin{table}[ht]
\centering
\renewcommand{\arraystretch}{0.95} 
\caption{Experiment 1: Average performance (\%) for single vs.\ multi-agent methods. Results cover seven key metrics, and bold indicates higher scores. Detailed per-drama breakdowns appear in Appendix~\ref{appendix:experiment_result}.}
\label{tab:comp_summary}
\begin{tabular}{l cc}
\hline
\textbf{Metric} & \textbf{Single} & \textbf{Multi} \\
\hline
Character Recall         & 54.1 & \textbf{72.9} \\
Group Match F1-Score     & 25.9 & \textbf{42.4} \\
Character-Relation Recall & 40.8 & \textbf{50.4} \\
Role Sim.          & 76.7 & \textbf{84.8} \\
Group Name Sim.    & 64.5 & \textbf{81.5} \\
Explicit Relation Sim.   & 70.2 & \textbf{74.0} \\
Implicit Relation Sim.   & 55.8 & \textbf{60.0} \\
\hline
\end{tabular}
\end{table}

%% file: tables/03_exp2.tex
\begin{table}[ht]
\centering
\renewcommand{\arraystretch}{0.95} 
\caption{
  Experiment~2: Average (\%) Precision, Recall, and F1 for PPR vs.\ Node-Edge Count. 
  Bolded values indicate higher performance. See Appendix~\ref{appendix:experiment_result2} for full per-drama details.
}
\label{tab:ppr_count_summary}
\begin{tabular}{l cc}
\hline
\textbf{Metric} & \textbf{PPR} & \textbf{Count} \\
\hline
Precision   & 46.8 & \textbf{47.2} \\
Recall      & 78.5 & \textbf{79.2} \\
F1 Score    & 58.2 & \textbf{58.7} \\
\hline
\end{tabular}
\end{table}

%% file: sections/08_conclusion.tex
\section{Conclusion}
This paper introduced \textbf{\ours{}}, a sequential LLM-based framework to automate and enhance character relationship extraction in long-form narratives. By chaining specialized LLM agents, \ours{} systematically tackles character composition, explicit/implicit relation detection, role identification, and group assignments. Experiments on Korean drama scripts confirm notable accuracy and completeness gains over single-agent baselines, underscoring \ours{}’s effectiveness for rapid, accurate script review. Nevertheless, challenges persist—particularly in robust group matching for evolving or ambiguous affiliations—highlighting the need for advanced inference techniques (e.g., external knowledge bases, deeper context modeling). Future extensions may include adaptive refinement strategies that align with shifting narratives and user-in-the-loop feedback to iteratively enhance \ours{}’s output. Overall, \ours{} holds substantial promise across entertainment, publishing, and educational domains, enabling efficient, high-quality narrative analysis while paving the way for further methodological refinements.

\section*{Acknowledgments}
We thank our colleagues at the AI R\&D Division for their valuable input and continued collaboration. We also thank the CJ ENM team for their support in evaluation and for providing constructive feedback throughout the development of this work.

%% file: sections/10_appendix.tex

\section{Prompt for Base Character Graph Construction}
\label{appendix:triplet_prompt}
\input{tables/10_triplet_generation_prompt}

\newpage
\section{Prompts for Multi-Agent Approach \ours} 
\label{appendix:prompt}

\subsection{Merging Duplicate Nodes}
\input{tables/05_multi_node_merge}

\newpage
\subsection{Relation Extraction}
\input{tables/06_1_multi_character_relation_extract}

\newpage
\input{tables/06_2_multi_character_relation_extract}

\newpage
\subsection{Filtering Out Irrelevant Characters}
\input{tables/07_multi_node_removal}

\newpage
\subsection{Role Identification}
\input{tables/08_multi_role_extraction}

\newpage
\subsection{Grouping Characters}
\input{tables/09_multi_node_grouping}

\twocolumn
\newpage
\section{Sample CRS Outputs at Each Refinement Step}
\label{appendix:stepbystep}

\input{figures/03_steps}
\onecolumn
\newpage
\section{Evaluation Metrics}
\label{appendix:metrics}
\input{tables/01_metrics}

\newpage
\section{Prompt for Single-Agent Approach (Experiment 1)}
\label{appendix:single_prompt}
\input{tables/04_singleprompt}


\newpage
\section{Experiment 1 Result}
\label{appendix:experiment_result}
\input{tables/02_exp1_appendix}
\newpage
\section{Experiment 2 Result}
\label{appendix:experiment_result2}
{\centering
\input{tables/03_exp2_appendix}
}


\twocolumn
\newpage
\section{Comparison of Character Extraction Algorithms}
\label{appendix:ppr_analysis_}
\input{figures/04_ppr}

%% file: tables/10_triplet_generation_prompt.tex
\begingroup 
\small 
\setlength{\tabcolsep}{4pt}
\begin{longtable}{p{0.15\linewidth}p{0.8\linewidth}}
\toprule
\textbf{System Prompt} & 
No system prompt was used. \\
\midrule
\textbf{User Prompt} & 
Read part of a story, then identify named entities and predicates. \newline

[Begin the drama script]\newline

\textbf{[Put your narrative chunk here]} \newline

[End the drama script] \newline
\\\bottomrule
\caption{Base Character Graph Construction Prompt.}
\label{tab:triplet_prompt}
\end{longtable}
\endgroup

%% file: tables/05_multi_node_merge.tex
\begingroup 
\small 
\setlength{\tabcolsep}{4pt}
\begin{longtable}{p{0.15\linewidth}p{0.8\linewidth}}
\toprule
\textbf{System Prompt} & 
No system prompt was used; all instructions were included within the user prompt. \\
\midrule
\textbf{User Prompt} & 
Your task is to identify any character pairs within the <Character Information> and <Summary> that refer to the same character but are listed under different names. Follow these specific guidelines:\newline

<Key Directives>\newline
- Examine character descriptions, relationships, and unique identifiers that could imply a single character referenced by multiple names (e.g., Seon-ae Eun - Hee-su's mother).\newline
- Use the <Character Information> and <Summary> for personality traits, roles, relationships, and affiliations to cross-check for consistent character depiction.\newline
- Each name set in a square bracket within <Character Name List> should be treated as a single identifier for a character. Do not split name sets with `/'.\newline
- When writing an answer, use the exact name set (e.g., `[Hee-su Park / Hee-su]') as it appears in <Character Name List>.\newline
- *** In Korean, the name `Young-min Cha' is composed of the last name `Cha' and the first name `Young-min'.\newline
- If two characters share the same first name, they are likely to be the same person. (`Young-min', `Young-min Cha')\newline
- If two characters share the same last name and one of their names includes an occupation title(`Young-min Cha' `Lawyer Cha'), they are also likely to be the same person. Be cautious of family members who may share a last name. Do not just assume they are the same person unless evidence supports it.\newline
<Source Priority>\newline
- If the length of <Character Information> is long enough, use <Character Information> as the primary source for grouping characters rather than <Summary>. \newline

<Provided Information>\newline
<Character Information> \textbf{[Put your treatment here]} </Character Information>\newline
<Summary> \textbf{[Put your summary here]} </Summary>\newline
<Character Name List> \textbf{[Put your all character list here]} </Character Name List>\newline
<Response Format>\newline
List any identified pairs of name sets from <Character Name List> referring to the same character as follows:
<Character Name List>\newline
<Character Pairs>\newline
1. **[Hee-su's mother]-[Seon-ae / Seon-ae Eun]** - Must list two different name sets from <Character Name List>. List only same person name pairs. Do not repeat the same name as answer name pairs. Do not include any partial name of name set as another name. If there is no same character within <Character Name List>, just write as [Character1]-[No Same Person].\newline
   **Evidence: [Describe the same identity of two names.]\newline
2. [Repeat for all characters from <Character Name List>.]\newline
</Response Format>\newline

<Output Example>\newline
<Character Name List Example>\newline
[Character A]
[Character B]
[Character C]
[Character D]
</Character Name List Example>\newline
Correct <Character Pairs> Example:\newline
1. **[Lee Ho]-[King]\newline
   **Evidence: King represents Lee Ho's title, and Lee Ho is the king's real name.**\newline
2. **[Hwang Gwi-in]-[No Same Person]\newline
3. **[Lawyer Hong] - [Hong Ji-yoon]: Lawyer Hong and Hong Ji-yoon are the same person.**\newline
Incorrect <Character Pairs> Example:\newline
1. **[Cheong-ha] - [Cheong-ha]\newline
2. **[Crown Prince] - [King]\newline
   **Evidence: The Crown Prince and the King are the same person, and King represents a title.\newline
Remember:\newline
- List names in the same order of <Character Name List>.\newline
- Do not repeat the same name as answer name pairs.\newline
- *** In Korean, the name `Young-min Cha' is composed of the last name `Cha' and the first name `Young-min'.\newline
- If two characters share the same first name, they are likely to be the same person.\newline
- If two characters share the same last name and one of their names includes an occupation title, they are also likely to be the same person. Be cautious of family members who may share a last name. Do not just assume they are the same person unless evidence supports it.
\\\bottomrule
\caption{Merging Duplicate Nodes Prompt.}
\label{tab:multi_agent_node_merge}
\end{longtable}
\endgroup

%% file: tables/06_1_multi_character_relation_extract.tex
\begingroup 
\small 
\setlength{\tabcolsep}{4pt}
\begin{longtable}{p{0.15\linewidth}p{0.8\linewidth}}
\toprule
\textbf{System Prompt} & 
No system prompt was used; all instructions were included within the user prompt. \\
\midrule
\textbf{User Prompt} & 
Your task is to identify and categorize explicit and implicit relationships for each character pair based on the \texttt{\textless Character Information\textgreater} and \texttt{\textless Summary\textgreater}. Relationships should not focus on temporal or incidental interactions but rather represent lasting or significant connections.\newline

<Key Directives>\newline
- Explicit Relationships:\newline
  - Limit explicit relationships to **romantic relationships**, **family relationships**, **friendships**, **teacher-student relationships** and **colleague relationships**. \newline
  - For romantic relationships, it can be ex-husband, ex-girlfriend and first-love and so on.\newline
  - For colleague relationships, specify whether the relationship is hierarchical (e.g., **supervisor**, **subordinate**) or equal (e.g., **colleague**, **peer**). Do not use a general term like `peer' for a hierarchical relationship.\newline
  - When characters share both initial relationships (e.g., medical school classmates, academy peers, etc.) and current professional roles (e.g., colleagues at the same hospital), prioritize the most defining and initial relationship (e.g., ``Peer'' for classmates) rather than the final one (e.g., ``colleague''). If the hierarchical relationship later becomes explicit (e.g., Supervisor/Subordinate), overriding the initial relationship.\newline
  - If the relationship is clearly stated in the \texttt{\textless Character Information\textgreater}, inferences based on the \texttt{\textless Summary\textgreater} is not needed.\newline
  - Clearly define family relationships (e.g., **father**, **daughter-in-law**, **Sibling**).\newline
  - For family relationships, ensure terms like `older brother', `younger sibling', and `older male cousin' are used only in their appropriate directional context. \newline
    For example:\newline
    `Hyeong (in Korean)' refers to an older male sibling or cousin, as seen from the perspective of the younger subject.\newline
    `Dongsaeng (in Korean)' refers to a younger sibling or cousin, as seen from the perspective of the older subject.\newline
  - Verify the subject-object relationship's directionality and correctness. If there is no explicit evidence supporting the relationship, write ``Information not provided.''\newline
  
- Implicit Relationships:\newline
  - Select the **most meaningful relationship** from the <Implicit Relationship List>.\newline 
  - Choose only relationship that reflects ongoing or significant dynamics throughout the \texttt{\textless Character Information\textgreater} and the \texttt{\textless Summary\textgreater}, not temporary or isolated incidents.\newline
  
- Relationship Extraction:\newline
  - Match the **exact subject-object pair order** as given in the \texttt{\textless Initial Character Knowledge Graph\textgreater}.\newline
  - If multiple significant relationships exist, pick the most relevant one.\newline
  - If no meaningful relationship exists, state "Information not provided" instead of guessing or adding irrelevant details.\newline
  
<Source Priority>\newline
- If the length of \texttt{\textless Character Information\textgreater} is long enough, use \texttt{\textless Character Information\textgreater} as the primary source for extracting relationships rather than \texttt{\textless Summary\textgreater}.\newline

<Implicit Relationship List>\newline
- Conflict, Betrayal, Affair, Help/Aid, Sacrifice, Dependency, Revenge, Resentment, Dislike, Worry/concern, One-sided love, Crush, Love, Longing, Love-hate relationship, Collaboration, Regret, Exploitation, Lie/Deception, Trust, Watching over/Protecting, Pressure, Conspiracy \newline
- Support: Use this term **only for power-related story**.\newline
- Friendliness: Use this term **only for power-related story**.\newline
- Hostility: Use this term **only for power-related story**.\newline
- Wariness: Use this term **when a character is cautious or suspicious towards another character, without full trust**.\newline

<Provided Information>\newline
<Character Information> \textbf{[Put your treatment here]} </Character Information>\newline
<Summary> \textbf{[Put your summary here]} </Summary>\newline
<Initial Character Knowledge Graph> \textbf{[Put your character pair list from the initial character graph here]} </Initial Character Knowledge Graph>

\\\bottomrule
\caption{Relation Extraction Prompt.}
\label{tab:multi_agent_relation_extraction}
\end{longtable}
\endgroup

%% file: tables/06_2_multi_character_relation_extract.tex
\begingroup 
\small 
\setlength{\tabcolsep}{4pt}
\addtocounter{table}{-1} 
\begin{longtable}{p{0.15\linewidth}p{0.8\linewidth}}
\toprule
\textbf{User Prompt} & 

<Response Format>\newline
1. **Subject: [As in Initial Character Knowledge Graph]**\newline
   **Object: [As in Initial Character Knowledge Graph]**\newline
   **(Explicit) Who is Subject regarding to Object]:** [If there is a hierarchical relationship between Subject and Object, use a specific term (e.g., supervisor, subordinate, advising professor) instead of a general one like `colleague'. If no meaningful relationship exists, state ``Information not provided''.]\newline
   **Verification: [Correct/Incorrect],** Verify that terms like older brother, younger sibling, older male cousin, and others are correctly applied based on the subject's perspective toward the object.\newline
   **(Implicit) What emotions does Subject experience toward Object?:** [Select from <Implicit Relationship List>. If no meaningful relationship exists, state ``Information not provided".]\newline
   
2. **Subject: [Repeat for the next pair]**\newline
   **Object: [Repeat for the next pair]**\newline
   **(Explicit) Who is Subject regarding to Object]:** [Repeat for the next pair]\newline
   **Verification: [Repeat for the next pair]**\newline
   **(Implicit) What emotions does Subject experience toward Object?:** [Repeat for the next pair]\newline
</Response Format>\newline

<Example>\newline
1. **Subject: Seung-tak Go / Seung-tak **\newline
   **Object: Seung-won / Han Seung-won **\newline
   **(Explicit) Who is Subject regarding to Object]: older male cousin**\newline
   **(Explicit) Verification: [Incorrect] Seung-tak Go is not Seung-won Han's older male cousin. Seung-won Han is Seung-tak Go's older male cousin.\newline
   **(Implicit) What emotions does Subject experience toward Object?: wariness**\newline
   
2. **Subject: Seung-won / Han Seung-won**\newline
   **Object: Seung-tak Go / Seung-tak**\newline
   **(Explicit) Who is Subject regarding to Object]: older male cousin**\newline
   **(Explicit) Verification: [Correct] Seung-won Han is Seung-tak Go's older male cousin.\newline
   **(Implicit) What emotions does Subject experience toward Object?: wariness**\newline

3. **Subject: Hyeri / Hyeri Go**\newline
   **Object: Sangwook / Sangwook Ju**\newline
   **(Explicit) Who is Subject regarding to Object]: colleague**\newline
   **(Explicit) Verification: [Incorrect] Hyeri is Sangwook's subordinate.\newline
   **(Implicit) What emotions does Subject experience toward Object?: wariness**\newline

3. **Subject: Sangwook / Sangwook Ju**\newline
   **Object: Hyeri / Hyeri Go**\newline
   **(Explicit) Who is Subject regarding to Object]: supervisor**\newline
   **(Explicit) Verification: [Correct] Sangwook is Hyeri's supervisor.\newline
   **(Implicit) What emotions does Subject experience toward Object?: wariness**\newline
   
</Example>\newline

***Key Reminders***:\newline
- Explicit relationships are limited to **family**, **friends**, and **colleagues**, with hierarchical roles specified for colleagues when relevant. **Do not use a general term like `colleague' for a hierarchical relationship, use a term as specific as possible. \newline
- Implicit relationships must be chosen strictly from the <Implicit Relationship List>. Use a checklist to ensure compliance.  \newline
- Relationships must be significant and non-temporal. Exclude incidental or temporary interactions.  \newline
- Do not change the subject-object pair order from the original <Initial Character Knowledge Graph>.  \newline
- For both explicit and implicit relationships, if no meaningful relationship exists, state ``Information not provided'' instead of guessing or adding irrelevant details.

\\\bottomrule
\caption{Relation Extraction Prompt. (Continued).}
\label{tab:multi_agent_relation_extraction}
\end{longtable}
\endgroup

%% file: tables/07_multi_node_removal.tex
\begingroup 
\small 
\setlength{\tabcolsep}{4pt}
\begin{longtable}{p{0.15\linewidth}p{0.8\linewidth}}
\toprule
\textbf{System Prompt} & 
No system prompt was used; all instructions were included within the user prompt. \\
\midrule
\textbf{User Prompt} & 
Your task is to identify 1) general character names, 2) inappropriate character identities, 3) inappropriate character relatioinships and 4) abundant relationships. Follow these specific guidelines:\newline
\newline
<Key Directives>\newline
1) Identify General Character Names:\newline
- Task: Review the \texttt{\textless Character List\textgreater}. Identify and select any names that seem to be a pronoun (e.g., \texttt{`doctor'}, \texttt{`patient'}, \texttt{`professor'}) not a specific character name. \newline
**If the last name exists it refers a specific person so do not include this type of name in the list (Manager Ban, Professor Cha, Sanggung Oh).** \newline
\newline
2) Identify Inappropriate Character Identities:\newline
- Task: Review the \texttt{\textless Character Identity List\textgreater}. Identify and select any entries that seem to describe an action, reaction, or state rather than a stable identity or occupation. \newline
\newline
3) Identify Inappropriate Character Relationships:\newline
- Task: Review the \texttt{\textless Character Relationship List\textgreater}. Identify and select entries that describe temporary actions or states instead of defining relationships. \newline
\newline
4) Identify Overused or Abundant Relationships:\newline
- Task: Review the \texttt{\textless Character Relationship List\textgreater} again. If any specific relationships occur too frequently and disrupt diversity in the relationships list, select these for review. \newline
\newline
<Provided Information>\newline
<Character List>\textbf{[Put your character list from previous step here]}</Character List>\newline
<Character Identity List>\textbf{[Put your matched character identity from initial graph here]}</Character Identity List>\newline
<Character Relationship List> \textbf{[Put your character pair list from initial graph here]} </Character Relationship List>\newline

<Response Format>\newline
1. General Character List:\newline
  1. **Character: [Write the exact name only from the \texttt{\textless Character List\textgreater}.]**\newline
    **Last Name: [True/False: Check if the name contains last name.]**\newline
  2. **[Continue to select all general character names from the \texttt{\textless Character List\textgreater}]**\newline
\newline
2. Inappropriate Character Identity List\newline
  **[Write an exact word from the \texttt{\textless Character Identity List\textgreater}]**\newline
  **[Continue to select all inappropriate character identities]**\newline
\newline
3. Inappropriate Character Relationship List\newline
  **[Write an exact word from the \texttt{\textless Character Relationship List\textgreater}]**\newline
  **[Continue to select all inappropriate character relationships]**\newline
\newline
4. Abundant Relationship List\newline
  **[Write an exact word from the \texttt{\textless Character Relationship List\textgreater}]**\newline
  **[Continue to select all abundant relationships]**\newline
  If there is no abundant relationship, state ``Information not provided''.\newline
\newline
</Response Format>\newline
\newline
Remember:\newline
- **If the last name exists it refers a specific person so do not include this type of name in General Character List.**
\\\bottomrule
\caption{Filtering Out Irrelevant Characters Prompt.}
\label{tab:multi_agent_node_removal}
\end{longtable}
\endgroup

%% file: tables/08_multi_role_extraction.tex
\begingroup 
\small 
\setlength{\tabcolsep}{4pt}
\begin{longtable}{p{0.15\linewidth}p{0.8\linewidth}}
\toprule
\textbf{System Prompt} & 
No system prompt was used; all instructions were included within the user prompt. \\
\midrule
\textbf{User Prompt} & 
Your task is to identify one main role (e.g., lawyer, cardiothoracic surgeon, unemployed person, king) for each character in \texttt{\textless Character Name Sets\textgreater} based on the \texttt{\textless Character Information\textgreater} and \texttt{\textless Summary\textgreater} provided. Relationships should be interpreted only in terms of each character’s primary profession, title, or distinctive role. \newline
\newline
<Key Directives> \newline
- Focus on extracting each character's primary profession, title, or distinctive role from the provided \texttt{\textless Character Information\textgreater} and \texttt{\textless Summary\textgreater}. \newline
- If there is a special role specified with the main role from \texttt{\textless Character Information\textgreater}, include them all separated by a comma (for example, ``group heir'', ``chairman Jang's son''). \newline
- **IMPORTANT: DO NOT interpret character names as roles, even if they sound descriptive. For example, a character named ``myeol-mang'' should not be assigned the role ``destroyer'' based on their name alone. \newline
- **DO NOT interpret relationships (e.g., `Yoon-gyeom Kang's wife') as the main role. \newline
- **Avoid general age terms like `adult', `elderly person', or `child', and do not include relationships. \newline
- Treat each name set within square brackets in \texttt{\textless Character Name Sets\textgreater} as a single identifier for a character. Do not split name sets based on `/'. \newline
- Use the exact name set (e.g., ``[Hee-su Park]'') as it appears in \texttt{\textless Character Name Sets\textgreater}. \newline
- If no clear role is identified from the provided information, write ``Information not provided.'' \newline
\newline
<Provided Information> \newline
<Character Information> \textbf{[Put your treatment here]} </Character Information>\newline
<Summary> \textbf{[Put your summary here]} </Summary>\newline
<Character Name Sets> \textbf{[Put your all character list here]} </Character Name Sets>\newline

<Response Format> \newline
List the primary role of each character in the \texttt{\textless Character Name Sets\textgreater}. If there are two primary roles, list them separated by a comma. \newline

1. **Character: [Do not alter any name set provided in \texttt{\textless Character Name Sets\textgreater}. Use an exact name set from \texttt{\textless Character Name Sets\textgreater}.]** \newline
  **Role: [Main role(s) of the character: Write with specific details if available. For example, instead of simply writing `prince', use `first prince', `second prince', or `third prince' if there are multiple siblings and this information is provided. Similarly, instead of simply writing `concubine', use `selected concubine' or `favored concubine' if the information is provided. Ensure roles reflect the character’s unique position or significance in the story context.]** \newline
  **Confidence: [High/Medium/Low] - Briefly explain the basis for this role assignment.** \newline
  
2. [Repeat as above for all name sets from \texttt{\textless Character Name Sets\textgreater}] \newline
<Examples> \newline
Correct: [lawyer Kim] - Role: prosecutor (based on information stating they work as a prosecutor) \newline
Incorrect: [lawyer Kim] - Role: lawyer (based solely on the character's name) \newline
\newline
<Source Priority> \newline
- If the length of \texttt{\textless Character Information\textgreater} is longer than a word, use \texttt{\textless Character Information\textgreater} as the primary source for extracting the character's role rather than \texttt{\textless Summary\textgreater}. \newline
\newline
Remember: \newline
- **Avoid general age terms like `adult', `elderly person', or `child', and do not include relationships.**
\\\bottomrule
\caption{Role Identification Prompt.}
\label{tab:multi_agent_character_role_extraction}
\end{longtable}
\endgroup

%% file: tables/09_multi_node_grouping.tex
\begingroup 
\small 
\setlength{\tabcolsep}{4pt}
\begin{longtable}{p{0.15\linewidth}p{0.8\linewidth}}
\toprule
\textbf{System Prompt} & 
No system prompt was used; all instructions were included within the user prompt. \\
\midrule
\textbf{User Prompt} & 
Your task is to identify and categorize essential groups of characters in <Character Role> based on family relationships, shared affiliations, professions, or significant roles, using the <Character Information>, <Summary>, and <Character List> provided. Follow these specific guidelines:\newline

<Key Directives>\newline
- Group characters exclusively based on their primary family membership, affiliation, profession, or role, ensuring each character is included in only one group. Each character should be assigned to the group that best reflects their most important function or connection within the story.\newline
- **IMPORTANT**: Group membership must be exclusive, meaning no character appears in more than one group. If a character fits multiple roles, place them in the group where their primary role is most relevant to the story context. If a character does not fit into any group, simply state ``No Group'' without attempting to assign them to one.\newline

<Grouping Priority>\newline
- **If the story is office-centric**, prioritize grouping characters by their affiliations, such as workplaces, departments, or professional roles. If characters belong to different departments within the same organization, create detailed and distinct groups for each department. For example, characters in the ``Public Relations (PR) team" should be grouped separately from those in the ``Planning team" even if they work for the same company.\newline
- **If the story is family-centric**, prioritize grouping characters by their family relationships. For instance, if a character is both a parent and a professional, they should be grouped based on their role as a parent.\newline
- Family group can be a very small group including mother and son.\newline

<Group Naming>\newline
- Use clear and specific names for each group that reflect characters’ shared roles or connections within the story context (e.g., ``Jurors'', ``Cardiothoracic staff'', ``Planning team''). \newline

<Provided Information>\newline
<Character Information> \textbf{[Put your treatment here]} </Character Information>\newline
<Summary> \textbf{[Put your summary here]} </Summary>\newline
<Character List> \textbf{[Put your all character list here]} </Character List>\newline

<Response Format>\newline
Use of <Character Information> as the primary source: True/False\newline
Do not include any other member out of <Character Name List> in the following response.\newline
List the essential groups identified, with exclusive members in each group.\newline
**Family Group List: [group\_A, group\_B group\_C, group\_D]**\newline
**Other Group List: [group\_A, group\_B group\_C, group\_D]**\newline
- Do not include generic group labels such as ``Others,'' ``Other characters,'' in the group list.\newline

<Character Name List>\newline
[List the character list from <Character List>]\newline

1. **Character: [**Use the exact name set from <Character List>. ]**\newline
  **Group: [Select one group from the Group List or state ``No Group''.]**\newline
  **Family: [If family members exists in <Character List>, answer as `Yes' otherwise `No'.]**\newline
  **Rationale: [Provide a reason why this character should belong to the Group.]\newline

2. [Repeat as above for all name sets from <Character List>]\newline

Remember:\newline
- **IMPORTANT**: Group membership is exclusive—no character should appear in more than one group.\newline
- **Use the exact name set from <Character List>, [Young-min / Young-min Cha] is the full name set.\newline
- **If the story focuses on family dynamics, prioritize grouping by family connections.** For example, if character1 is the child of character2 and character2's identity/affiliation is less important than their role as character1's parent, group them together based on the family relationship.\newline
- **If the story is office-focused, prioritize grouping by affiliations or professional roles.**
\\\bottomrule
\caption{Grouping Characters Prompt.}
\label{tab:multi_agent_node_grouping}
\end{longtable}
\endgroup

%% file: figures/03_steps.tex
\begin{figure*}[htbp]
    \centering
    \begin{subfigure}[b]{0.45\textwidth}
        \centering
        \includegraphics[width=\textwidth, page=1]{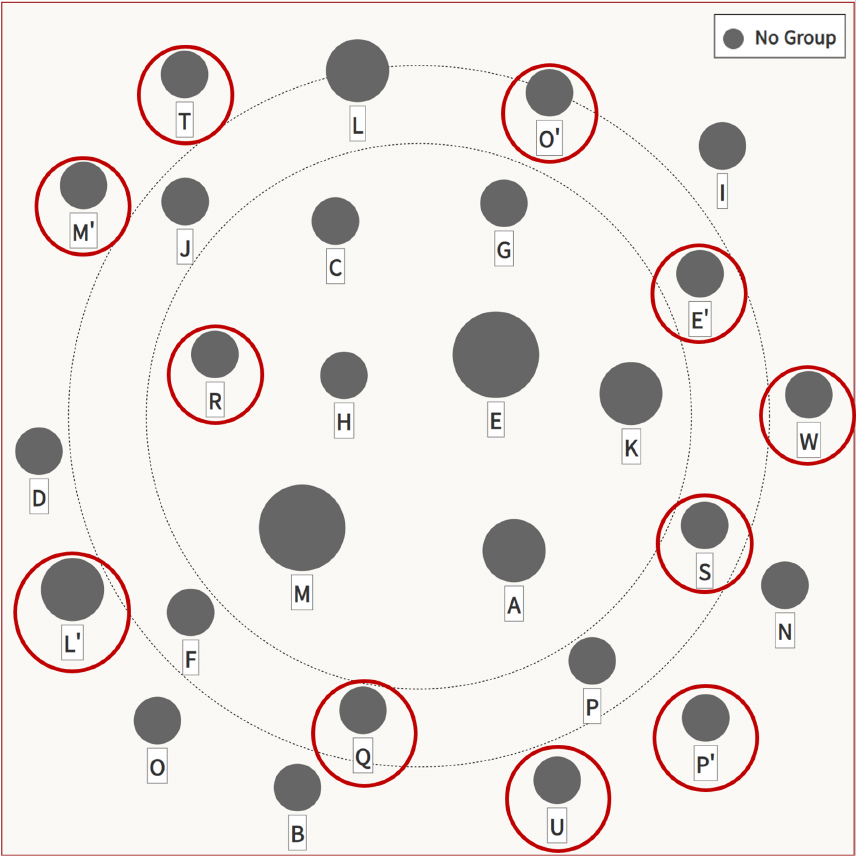}
        \caption{\textbf{Step 1}: After PPR-based Character Selection. Red circles = not in GT.}
    \end{subfigure}
    \hspace{3mm}
    \begin{subfigure}[b]{0.45\textwidth}
        \includegraphics[width=\textwidth, page=1]{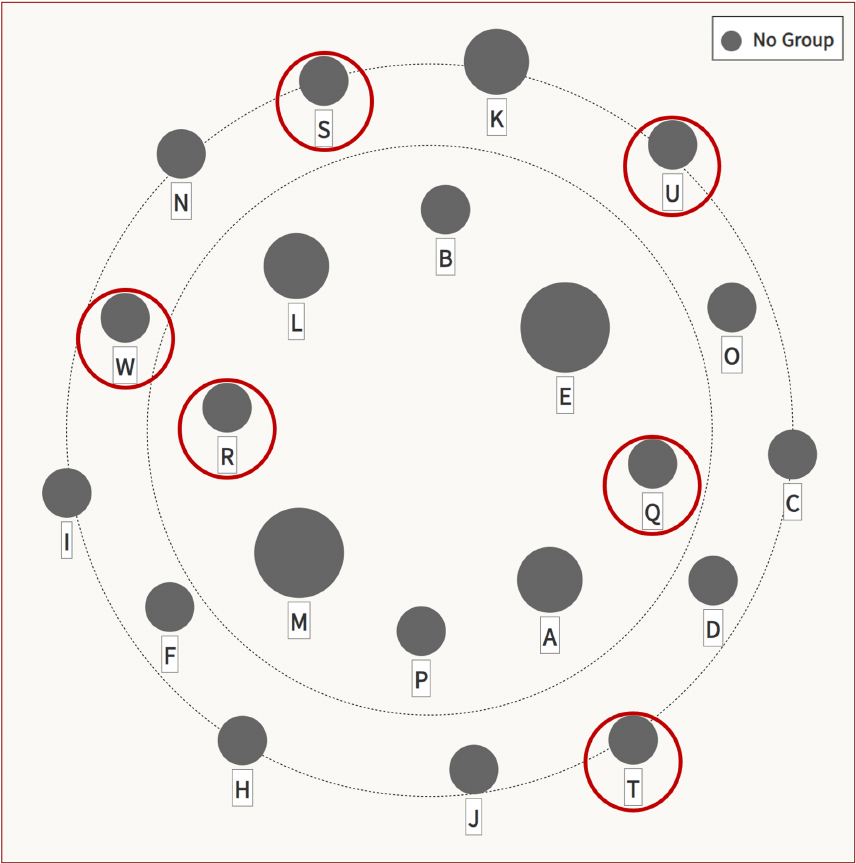}
        \caption{\textbf{Step 2}: After Node Merge. Duplicate nodes (E', L', M', O', P') removed.}
    \end{subfigure}
    \begin{subfigure}[b]{0.45\textwidth}
        \includegraphics[width=\textwidth, page=1]{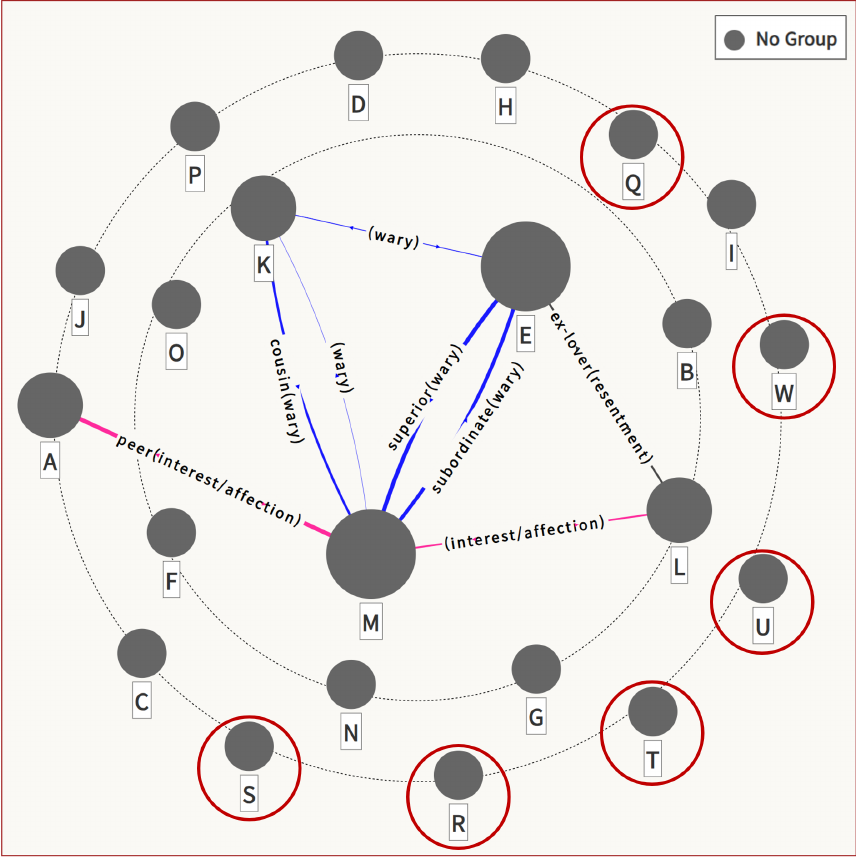}
        \caption{\textbf{Step 3}: After Relation Extraction. Labels added for main/sub-character links.}
    \end{subfigure}
    \hspace{3mm}
    \begin{subfigure}[b]{0.45\textwidth}
        \includegraphics[width=\textwidth, page=1]{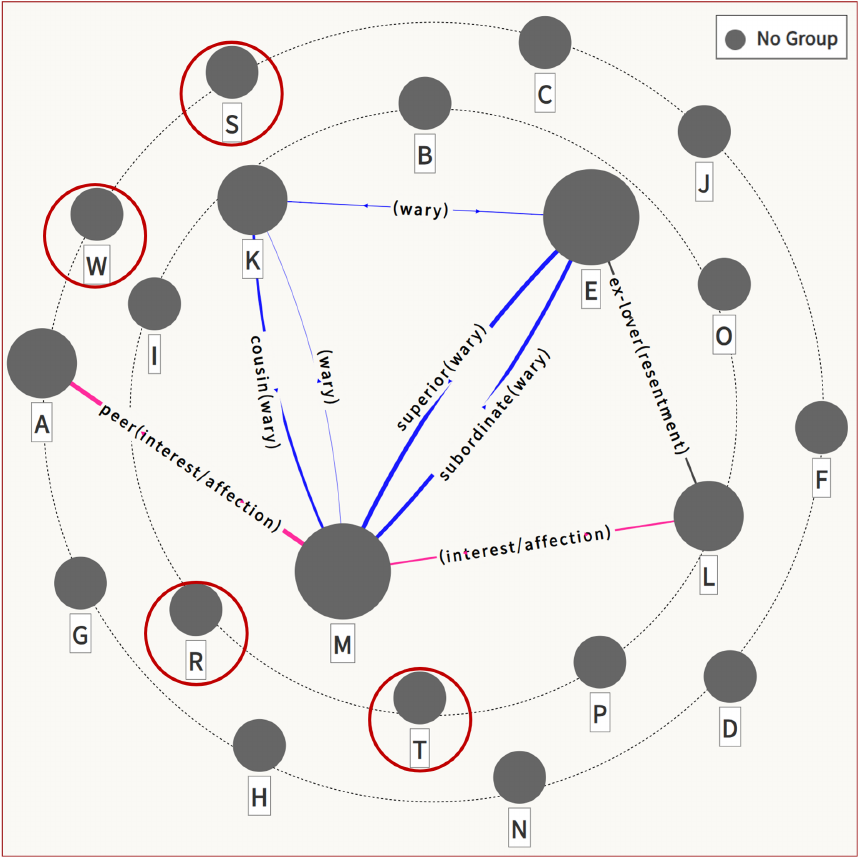}
        \caption{\textbf{Step 4}: After Filtering Out. Irrelevant characters (e.g., U, Q) discarded.}
    \end{subfigure}
    \begin{subfigure}[b]{0.45\textwidth}
        \includegraphics[width=\textwidth, page=1]{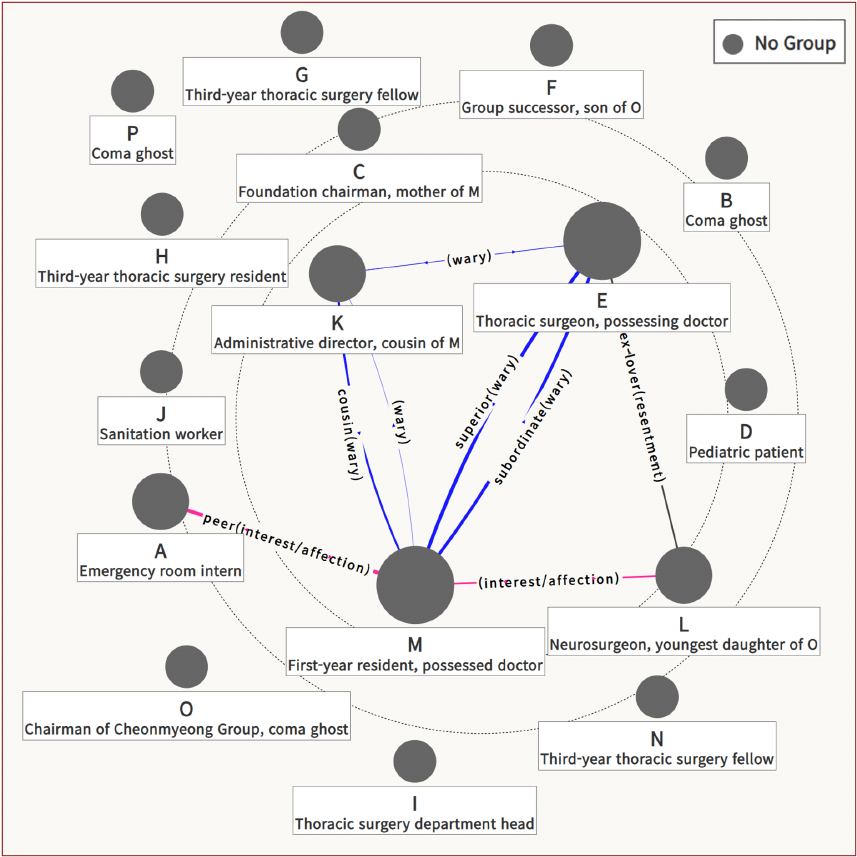}
        \caption{\textbf{Step 5}: After Role Identification. Roles assigned.}
    \end{subfigure}
    \hspace{3mm}
    \begin{subfigure}[b]{0.45\textwidth}
        \includegraphics[width=\textwidth, page=1]{figures/Ghost_Doctor_p.pdf}
        \caption{\textbf{Step 6}: Final CRS following Grouping.}
    \end{subfigure}
    
    \caption{Step-by-step refinement of the CRS for \textbf{Drama~13}. Each subfigure illustrates the incremental changes at a distinct process stage (see Section~\ref{ssec:seq_refinement} for details).}
    \label{fig:stepbystep}
\end{figure*}

Figure~\ref{fig:stepbystep} illustrates how the CRS for Drama~13 evolves at each refinement stage. The largest circles indicate main characters, medium circles represent sub-characters, and smaller circles denote supporting or minor characters.

\begin{itemize}
    \item \textbf{(a) Step 1---PPR-Based Character Selection:} 
    An initial list of characters is chosen using PPR. Red nodes correspond to individuals not appearing in the ground truth (GT).
    \item \textbf{(b) Step 2---Node Merge:} 
    Duplicate or aliased nodes (e.g., E', L', M', O', P') are consolidated into single entities, reducing redundancy.
    \item \textbf{(c) Step 3---Relation Extraction:} 
    Main and sub-character relations are labeled (explicitly or implicitly).
    Pink edges signify ``affectionate'' connections, while blue edges denote ``cautious'' interactions.
    \item \textbf{(d) Step 4---Filtering Out:} 
    Characters deemed irrelevant, such as U and Q, are removed to simplify the graph and focus on key roles.
    \item \textbf{(e) Step 5---Role Identification:} 
    Character roles (e.g., ‘doctor,’ ‘student’) are assigned. Nodes lacking definable roles (S, T, R, W) are eliminated to maintain clarity.
    \item \textbf{(f) Step 6---Grouping Characters:} 
    The final CRS clusters characters by group affiliation, with each group distinguished by a unique color or label. Group names are displayed in the figure's top-right corner.
\end{itemize}

%% file: tables/01_metrics.tex
\begin{table*}[htbp]
    \centering
    \small
    \resizebox{\linewidth}{!}{ 
    \begin{tabular}{lp{10cm}}
        \toprule
        \textbf{Metric} & \textbf{Definition} \\
        \midrule
        \textbf{Character Recall} & Number of matched characters between ground truth and predicted data / Total number of characters in the ground truth \\
        \midrule
        \textbf{Role Similarity} & Sum of similarity scores between roles of matched characters in ground truth and predictions / Number of matched characters \\
        \midrule
        \textbf{Group Match F1-score} & TP (True Positive): Character pairs that belong to the same group in both ground truth and predictions \newline
        FP (False Positive): Character pairs grouped together in predictions but not in ground truth \newline
        FN (False Negative): Character pairs grouped together in ground truth but not in predictions \newline
        TN (True Negative): Character pairs belonging to different groups in both ground truth and predictions \newline
        
        \textbf{F1-score} :  
        \[
        \text{Precision} = \frac{TP}{TP + FP}, \quad
        \text{Recall} = \frac{TP}{TP + FN}
        \]
        \[
        \text{F1-score} = \frac{2 \times \text{Precision} \times \text{Recall}}{\text{Precision} + \text{Recall}}
        \] \\
        \midrule
        \textbf{Group Name Similarity} & Sum of similarity scores between predicted and ground truth group names for True Positive characters / Total number of True Positive characters \\
        \midrule
        \textbf{Character-Relation Recall} & Number of matched character pairs between ground truth and predictions / Total number of character pairs in ground truth \\
        \midrule
        \textbf{Explicit Relation Similarity} & Sum of similarity scores between explicit relations of matched character pairs / Number of matched character pairs \\
        \midrule
        \textbf{Implicit Relation Similarity} & Sum of similarity scores between implicit relations of matched character pairs / Number of matched character pairs \\
        \bottomrule
    \end{tabular}
    }
    \caption{Definition of Evaluation Metrics.}
    \label{tab:metrics}
\end{table*}

%% file: tables/04_singleprompt.tex
\begingroup 
\small 
\setlength{\tabcolsep}{4pt}
\begin{longtable}{p{0.15\linewidth}p{0.8\linewidth}}
\toprule
\textbf{System Prompt} & 
No system prompt was used; all instructions were included within the user prompt. \\
\midrule
\textbf{User Prompt} & 
1. Your task is to identify any character pairs within the <Character Information> and <Summary> that refer to the same character but are listed under different names. \newline
2. Your task is to identify and categorize explicit and implicit relationships for each character pair based on the <Character Information> and <Summary>. Relationships should not focus on temporal or incidental interactions but rather represent lasting or significant connections. \newline
3. Your task is to identify general character names. \newline
4. Your task is to identify one main role (e.g., lawyer, cardiothoracic surgeon, unemployed, king) for each character in <Character Name Sets> based on the <Character Information> and <Summary> provided. \newline
5. Your task is to identify and categorize essential groups of characters in <Character Role> based on family relationships, shared affiliations, professions, or significant roles, using the <Character Information>, <Summary>, and <Character List> provided. \newline
6. Your task is to to leverage diverse perspectives to arrive at the most robust and accurate character relationship chart for the K-drama, considering the factor (**Power dynamics and alliances**, **Family relationships**, **Romantic relationships**) \newline

<Key Directives for task \#1> 
\textbf{[Key Directives used for node merge prompt (Table \ref{tab:multi_agent_node_merge})]}

<Key Directives for task \#2> 
\textbf{[Key Directives used for relation extraction (Table \ref{tab:multi_agent_relation_extraction})]}

<Key Directives for task \#3> 
\textbf{[Key Directives used for node removal prompt (Table \ref{tab:multi_agent_node_removal})]}

<Key Directives for task \#4> 
\textbf{[Key Directives used for role extraction prompt (Table \ref{tab:multi_agent_character_role_extraction})]}

<Key Directives for task \#5> 
\textbf{[Key Directives used for node grouping prompt (Table \ref{tab:multi_agent_node_grouping})]}\newline

<Source Priority> \newline
- If the length of <Character Information> is long enough, use <Character Information> as the primary source for grouping characters rather than <Summary>. \newline

<Provided Information> \newline
<Character Information> \textbf{[Put your treatment here]} </Character Information> \newline
<Summary> \textbf{[Put your summary here]} </Summary> \newline
<Character Name List> \textbf{[Put your all character list here]} </Character Name List> \newline
<Main-Sub Character List> \textbf{[Put your main and sub character list here]} </Main-Sub Character List> \newline

<Output Example> \newline
\#\#\# Rules for Relationship Entries: \newline
1. When Subject and Object are the same person: \newline
**Subject Character Name** : **Name1** (in Korean) \newline
**Object Character Name** : **Name2** (in Korean) \newline
**Explicit Relation**: **None** (When Subject and Object are the same person!) \newline
**Implicit Relation**: **None** (When Subject and Object are the same person!) \newline
**Role**: Fill in the character’s role  \newline
**Group**: Fill in the character’s group \newline
2. If Subject and Object are NOT the same person, AND both are in Main-Sub Character list: \newline
**Subject Character Name** : **Name1** (in Korean) \newline
**Object Character Name** : **Name2** (in Korean) \newline
**Explicit Relation**: Describe the superficial relationship (e.g., `Sibling', `Mentor', `Rival'). \newline
**Implicit Relation**: Choose a hidden or indirect relationship from <Implicit Relationship List>. \newline
**Role**: **None** (If Subject and Object are NOT the same person!) \newline
**Group**: **None** (If Subject and Object are NOT the same person!) \newline
3.If either Subject or Object is NOT a Main-Sub Character, exclude them \newline

List of All Characters from <Character Name List> excluding general names and duplicates. \newline
\#\#\# List of Characters: \newline
**[Name1]** \newline
**[Name2]** \newline

Provide a comprehensive answer for all tasks at once in the following Relationship Data Format. Do not write each output for all tasks. \newline
</Output Example> \newline

\\\bottomrule
\caption{Single Agent Prompt.}
\label{tab:single_agent}
\end{longtable}
\endgroup

%% file: tables/02_exp1_appendix.tex
\begin{table}[htbp]
\centering
\resizebox{0.8\textwidth}{!}{%
\begin{tabular}{l|cc|cc|cc}
\hline
\multirow{2}{*}{\textbf{Drama ID (Year)}} 
& \multicolumn{2}{c|}{\textbf{Character Recall}} 
& \multicolumn{2}{c|}{\textbf{Group Match F1-Score}} 
& \multicolumn{2}{c}{\textbf{Character-Relation Recall}} \\
\cline{2-7}
& \textbf{Single} & \textbf{Multi} 
& \textbf{Single} & \textbf{Multi} 
& \textbf{Single} & \textbf{Multi} \\
\hline
\textbf{Drama 1 (2014)}  & 60.0 & \textbf{86.7}  & \textbf{35.7} & 0.0  & 6.3  & \textbf{31.3} \\
\textbf{Drama 2 (2014)}  & \textbf{72.2} & 55.6  & \textbf{43.8} & 40.0  & 23.5  & \textbf{82.4} \\
\textbf{Drama 3 (2015)}  & 60.0 & \textbf{73.3}  & \textbf{38.7} & 29.6  & 30.8  & \textbf{69.2} \\
\textbf{Drama 4 (2016)}  & 55.0 & \textbf{60.0}  & 5.6 & \textbf{32.4}  & \textbf{41.2} & 23.5 \\
\textbf{Drama 5 (2017)}  & 69.2 & 69.2  & \textbf{35.7} & 33.3  & 11.1  & \textbf{27.8} \\
\textbf{Drama 6 (2018)}  & 55.0 & \textbf{75.0}  & 30.1 & \textbf{61.8}  & 54.6  & \textbf{68.2} \\
\textbf{Drama 7 (2018)}  & 68.8 & \textbf{87.5}  & 29.5 & \textbf{46.2}  & 0.0   & \textbf{24.1} \\
\textbf{Drama 8 (2018)}  & \textbf{44.4} & 38.9  & 22.6 & \textbf{37.0}  & \textbf{53.3} & 33.3 \\
\textbf{Drama 9 (2018)}  & 0.0  & \textbf{53.3}  & 0.0 & \textbf{27.6}  & \textbf{100.0} & 84.6 \\
\textbf{Drama 10 (2022)} & 56.0 & \textbf{76.0}  & 13.3 & \textbf{34.4}  & 13.0  & \textbf{21.7} \\
\textbf{Drama 11 (2018)} & 34.8 & \textbf{73.9}  & 9.1  & \textbf{49.4}  & \textbf{75.0} & 58.3\\
\textbf{Drama 12 (2022)} & 40.0 & \textbf{100.0} & 18.2& \textbf{82.7}  & 47.1  & 47.1 \\
\textbf{Drama 13 (2022)} & 87.5 & \textbf{100.0} & 55.6 & \textbf{90.5}  & 62.5  & \textbf{100.0} \\
\textbf{Drama 14 (2022)} & 17.7 & \textbf{70.6}  & 0.0  & \textbf{16.7}  & 71.4  & 71.4 \\
\textbf{Drama 15 (2023)} & \textbf{90.9} & 72.7  & 51.2 & \textbf{53.6}  & \textbf{21.7} & 13.0 \\
\hline
\textbf{Average}       & 54.1 \scriptsize(± 23.4) & \textbf{72.9} \scriptsize(± 16.2)  & 25.9 \scriptsize(± 17.3) & \textbf{42.4} \scriptsize(± 22.5)  & 40.8\scriptsize(± 28.0) & \textbf{50.4} \scriptsize(± 26.6) \\
\hline
\end{tabular}%
}
\caption{Experiment 1 (Part A): Comparison of single-agent vs.\ multi-agent methods (Charater Recall, Group Match F1-score, Character-Relation Recall)  in \%.}
\label{tab:comp_main_subset1}
\end{table}

\begin{table}[htbp]
\centering
\resizebox{\textwidth}{!}{%
\begin{tabular}{l|cc|cc|cc|cc}
\hline
\multirow{2}{*}{\textbf{Drama ID (Year)}} 
& \multicolumn{2}{c|}{\textbf{Role Sim.}} 
& \multicolumn{2}{c|}{\textbf{Group Name Sim.}} 
& \multicolumn{2}{c|}{\textbf{Explicit Relation Sim.}} 
& \multicolumn{2}{c}{\textbf{Implicit Relation Sim.}} \\
\cline{2-9}
& \textbf{Single} & \textbf{Multi} 
& \textbf{Single} & \textbf{Multi} 
& \textbf{Single} & \textbf{Multi} 
& \textbf{Single} & \textbf{Multi} \\
\hline
\textbf{Drama 1 (2014)}  & 87.7 \scriptsize(± 14.0) & \textbf{89.1} \scriptsize(± 11.5)  & \textbf{61.7} \scriptsize(± 5.6)  & 0.0 \scriptsize(± 0.0)  & \textbf{100.0} \scriptsize(± 0.0)  & 64.1 \scriptsize(± 2.3)  & \textbf{65.4} \scriptsize(± 0.0)  & 64.2 \scriptsize(± 4.1) \\
\textbf{Drama 2 (2014)}  & \textbf{84.4} \scriptsize(± 12.6) & 83.9 \scriptsize(± 13.9)  & 87.8 \scriptsize(± 17.3)  & \textbf{100.0} \scriptsize(± 0.0)  & 74.7 \scriptsize(± 7.7)  & \textbf{75.1} \scriptsize(± 7.8)  & \textbf{63.9} \scriptsize(± 2.4)  & 61.8 \scriptsize(± 3.1) \\
\textbf{Drama 3 (2015)}  & 81.6 \scriptsize(± 12.8) & \textbf{88.5} \scriptsize(± 11.2)  & 86.6 \scriptsize(± 16.3)  & \textbf{87.5} \scriptsize(± 14.3)  & \textbf{72.2} \scriptsize(± 3.5)  & 60.8 \scriptsize(± 35.0)  & \textbf{63.5} \scriptsize(± 5.2)  & 40.5 \scriptsize(± 34.9) \\
\textbf{Drama 4 (2016)}  & \textbf{86.3} \scriptsize(± 13.5) & 78.4 \scriptsize(± 18.7)  & 64.4 \scriptsize(± 0.0)  & \textbf{87.2} \scriptsize(± 0.0)  & \textbf{66.2} \scriptsize(± 30.3)  & 56.0 \scriptsize(± 32.7)  & 38.2 \scriptsize(± 33.2)  & \textbf{71.1} \scriptsize(± 16.8) \\
\textbf{Drama 5 (2017)}  & 79.1 \scriptsize(± 12.0) & \textbf{80.3} \scriptsize(± 12.7)  & \textbf{74.8} \scriptsize(± 15.4)  & 70.5 \scriptsize(± 12.7)  & \textbf{100.0} \scriptsize(± 0.0)  & 73.2 \scriptsize(± 5.6)  & \textbf{81.1} \scriptsize(± 18.9)  & 76.1 \scriptsize(± 19.5) \\
\textbf{Drama 6 (2018)}  & \textbf{73.8} \scriptsize(± 14.0) & 69.1 \scriptsize(± 12.5)  & 62.4 \scriptsize(± 2.6)  & \textbf{78.1} \scriptsize(± 9.6)  & \textbf{76.2} \scriptsize(± 14.0)  & 65.6 \scriptsize(± 18.0)  & \textbf{71.5} \scriptsize(± 16.8)  & 65.2 \scriptsize(± 9.8) \\
\textbf{Drama 7 (2018)}  & 77.9 \scriptsize(± 13.9) & \textbf{84.3} \scriptsize(± 13.4)  & 87.5 \scriptsize(± 5.3)  & \textbf{92.3} \scriptsize(± 1.8)  & 0.0 \scriptsize(± 0.0)  & \textbf{90.6} \scriptsize(± 11.0)  & 0.0 \scriptsize(± 0.0)  & \textbf{65.8} \scriptsize(± 3.9) \\
\textbf{Drama 8 (2018)}  & \textbf{91.0} \scriptsize(± 10.0) & 79.6 \scriptsize(± 11.9)  & 58.4 \scriptsize(± 0.0)  & \textbf{76.3} \scriptsize(± 0.0)  & \textbf{82.0} \scriptsize(± 14.2)  & 68.3 \scriptsize(± 36.7)  & 67.0 \scriptsize(± 7.8)  & \textbf{69.2} \scriptsize(± 7.8) \\
\textbf{Drama 9 (2018)}  & 0.0 \scriptsize(± 0.0)  & \textbf{93.4} \scriptsize(± 14.0)  & 0.0 \scriptsize(± 0.0)  & \textbf{85.1} \scriptsize(± 0.6)  & \textbf{59.5} \scriptsize(± 41.9)  & 55.0 \scriptsize(± 42.7)  & 20.8 \scriptsize(± 31.3)  & \textbf{62.1} \scriptsize(± 3.6) \\
\textbf{Drama 10 (2022)} & 85.0 \scriptsize(± 10.8) & \textbf{90.5} \scriptsize(± 11.9)  & \textbf{97.3} \scriptsize(± 0.2)  & 97.1 \scriptsize(± 0.8)  & \textbf{94.8} \scriptsize(± 3.9)  & 84.8 \scriptsize(± 12.5)  & 58.0 \scriptsize(± 2.3)  & \textbf{65.0} \scriptsize(± 2.7) \\
\textbf{Drama 11 (2018)} & 81.8 \scriptsize(± 10.5) & \textbf{89.8} \scriptsize(± 7.2)  & 65.5 \scriptsize(± 8.1)  & \textbf{83.5} \scriptsize(± 15.2)  & 50.08 \scriptsize(± 45.8)  & \textbf{93.1} \scriptsize(± 11.1)  & 68.0 \scriptsize(± 29.6)  & \textbf{76.5} \scriptsize(± 16.5) \\
\textbf{Drama 12 (2022)} & 74.9 \scriptsize(± 16.0) & \textbf{88.3} \scriptsize(± 15.1)  & 59.9 \scriptsize(± 3.0)  & \textbf{100.0} \scriptsize(± 0.0)  & 77.8 \scriptsize(± 15.8)  & \textbf{100.0} \scriptsize(± 0.0)  & \textbf{32.8} \scriptsize(± 43.7)  & 0.0 \scriptsize(± 0.0) \\
\textbf{Drama 13 (2022)} & 96.2 \scriptsize(± 10.4) & \textbf{96.8} \scriptsize(± 6.8)  & 100.0 \scriptsize(± 0.0)  & 100.0 \scriptsize(± 0.0)  & 73.4 \scriptsize(± 7.5)  & \textbf{85.8} \scriptsize(± 32.7)  & \textbf{84.4} \scriptsize(± 19.2)  & 80.4 \scriptsize(± 16.1) \\
\textbf{Drama 14 (2022)} & 57.7 \scriptsize(± 4.1)  & \textbf{69.7} \scriptsize(± 26.7)  & 0.0 \scriptsize(± 0.0)  & \textbf{100.0} \scriptsize(± 0.0)  & 57.8 \scriptsize(± 30.6)  & \textbf{62.9} \scriptsize(± 23.4)  & \textbf{70.1} \scriptsize(± 16.0)  & 58.6 \scriptsize(± 20.9) \\
\textbf{Drama 15 (2023)} & \textbf{93.8} \scriptsize(± 7.4)  & 89.9 \scriptsize(± 13.0)  & 61.0 \scriptsize(± 2.6)  & \textbf{64.2} \scriptsize(± 3.5)  & 68.0 \scriptsize(± 7.1)  & \textbf{74.1} \scriptsize(± 6.8)  & \textbf{52.8} \scriptsize(± 27.3)  & 44.3 \scriptsize(± 31.3) \\
\hline
\textbf{Average}       & 76.7 \scriptsize(± 22.4) & \textbf{84.8} \scriptsize(± 7.8)  & 64.5 \scriptsize(± 28.8) & \textbf{81.5} \scriptsize(± 24.4)  & 70.2 \scriptsize(± 23.5) & \textbf{74.0} \scriptsize(± 13.6)  & 55.8 \scriptsize(± 22.5) & \textbf{60.0} \scriptsize(± 19.1) \\
\hline
\end{tabular}%
}
\caption{Experiment 1 (Part B): Comparison of single-agent vs.\ multi-agent methods (Role Similarity, Group Name Similarity, Explicit Relation Similarity, Implicit Relation Similarity) in \%.}
\label{tab:comp_main_subset2}
\end{table}

%% file: tables/03_exp2_appendix.tex
\FloatBarrier
\begin{table*}[htbp]
    \centering
    \resizebox{0.7\textwidth}{!}{%
    \begin{tabular}{lcc|cc|cc}
    \hline
    \multirow{2}{*}{\textbf{Drama ID (Year)}} 
    & \multicolumn{2}{c|}{\textbf{Precision}} 
    & \multicolumn{2}{c|}{\textbf{Recall}} 
    & \multicolumn{2}{c}{\textbf{F1 Score}} \\
    \cline{2-7}
    & \textbf{PPR} & \textbf{Count} 
    & \textbf{PPR} & \textbf{Count} 
    & \textbf{PPR} & \textbf{Count} \\
    \hline
    \textbf{Drama 1 (2014)}  
    & 44.1 & 44.1 
    & 100.0 & 100.0
    & 61.2 & 61.2 \\
    \textbf{Drama 2 (2014)}  
    & \textbf{50.0} & 45.8 
    & \textbf{66.7} & 61.1 
    & \textbf{57.1} & 52.4 \\
    \textbf{Drama 3 (2015)}  
    & 38.7 & \textbf{41.9} 
    & 80.0 & \textbf{86.7} 
    & 52.2 & \textbf{56.5} \\
    \textbf{Drama 4 (2016)}  
    & 51.6 & \textbf{61.3} 
    & 80.0 & \textbf{95.0} 
    & 62.8 & \textbf{74.5} \\
    \textbf{Drama 5 (2017)}  
    & 32.1 & 32.1
    & 69.2 & 69.2
    & 43.9 & 43.9 \\
    \textbf{Drama 6 (2018)}  
    & 45.7 & 45.7 
    & 80.0 & 80.0 
    & 58.2 & 58.2 \\
    \textbf{Drama 7 (2018)}  
    & 37.8 & 37.8
    & 87.5 & 87.5 
    & 52.8 & 52.8 \\
    \textbf{Drama 8 (2018)}  
    & 25.9 & \textbf{29.6} 
    & 38.9 & \textbf{44.4} 
    & 31.1 & \textbf{35.6} \\
    \textbf{Drama 9 (2018)}  
    & \textbf{30.0} & 26.7
    & \textbf{60.0} & 53.3
    & \textbf{40.0} & 35.6 \\
    \textbf{Drama 10 (2022)} 
    & 54.8 & \textbf{71.0} 
    & 73.9 & \textbf{95.7} 
    & 63.0 & \textbf{81.5} \\
    \textbf{Drama 11 (2018)} 
    & 57.1 & 57.1 
    & 80.0 & 80.0 
    & 66.7 & 66.7 \\
    \textbf{Drama 12 (2022)} 
    & \textbf{71.4} & 65.7 
    & \textbf{100.0} & 92.0 
    & \textbf{83.3} & 76.7 \\
    \textbf{Drama 13 (2022)} 
    & \textbf{59.3} & 55.6 
    & \textbf{100.0} & 93.8 
    & \textbf{74.4} & 69.8 \\
    \textbf{Drama 14 (2022)} 
    & 51.7 & \textbf{55.2} 
    & 88.2 & \textbf{94.1} 
    & 65.2 & \textbf{69.6} \\
    \textbf{Drama 15 (2023)} 
    & \textbf{51.6} & 38.7 
    & \textbf{72.7} & 54.6 
    & \textbf{60.4} & 45.3 \\
    \hline
    \textbf{Average} 
    & 46.8 \scriptsize(± 11.8) & \textbf{47.2} \scriptsize(± 12.9)
    & 78.5 \scriptsize(± 15.8) & \textbf{79.2} \scriptsize(± 17.5)
    & 58.15 \scriptsize(± 12.7) & \textbf{58.7} \scriptsize(± 14.0) \\
    \hline
    \end{tabular}%
    }
    \caption{Experiment 2: Performance comparison of character selection (PPR vs.\ Node-Edge Count) in \%.}
    \label{tab:metrics_table2}
\end{table*}
\FloatBarrier

%% file: figures/04_ppr.tex

\begin{minipage}{\textwidth}
    \centering
    \includegraphics[width=\textwidth]{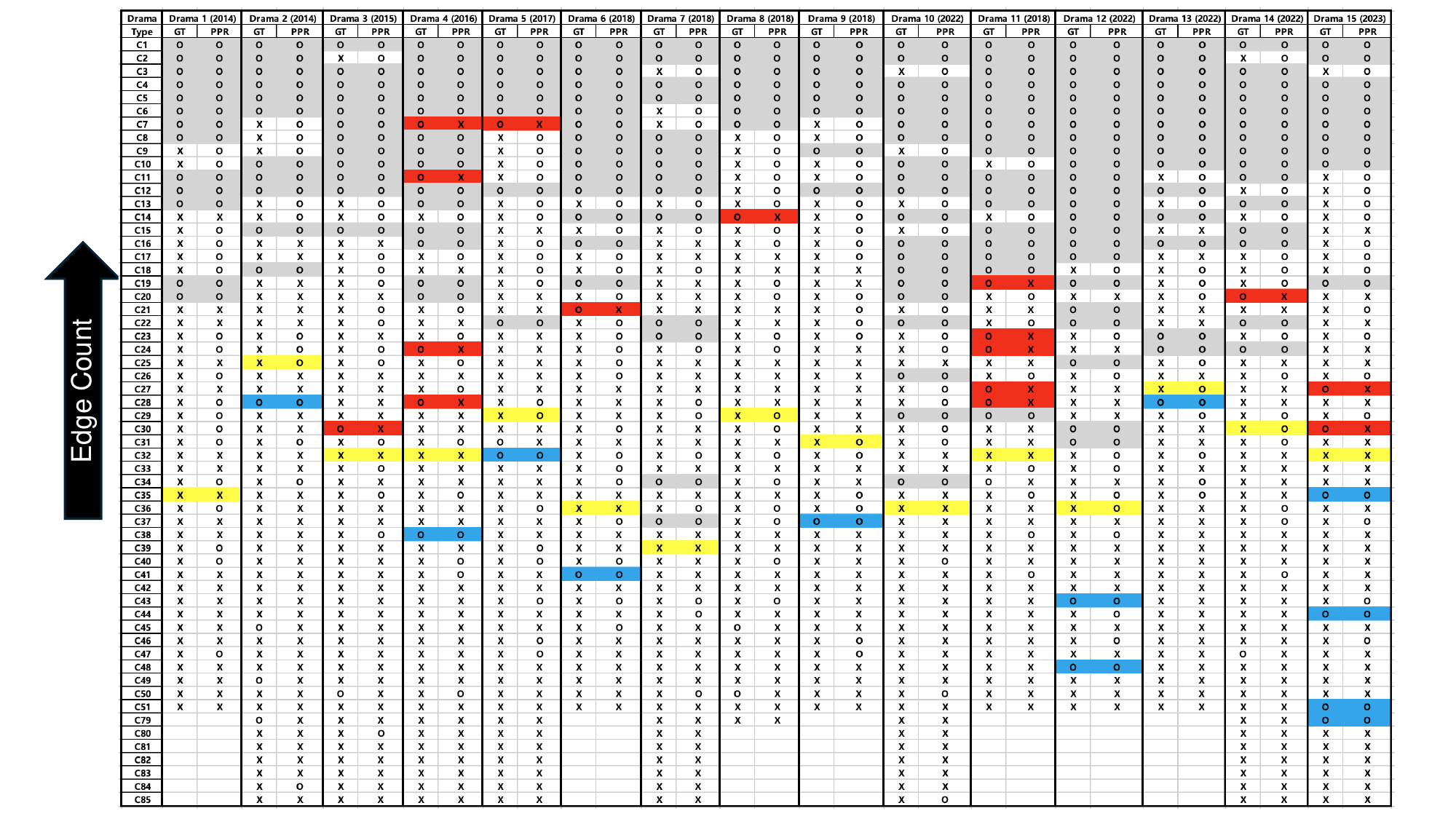}
    \captionof{figure}{Comparison of Character Extraction Algorithm (PPR vs.\ Edge Count).}
    \label{ppr_analysis}
\end{minipage}

\vspace{1em}

Figure~\ref{ppr_analysis} displays a color-coded table comparing the GT characters against those extracted by both the PPR algorithm and an edge-count method. Characters are sorted by descending edge count (C1 = highest). The yellow-highlighted region marks the cutoff determined by the edge-count approach, matching the total characters selected by PPR. Gray cells indicate cases where both methods align with the GT, blue cells indicate where PPR uniquely succeeds, and red cells highlight scenarios where the edge-count method outperforms PPR. Notably, in dramas~2, 9, 12, 13, and~15—where key GT characters appear less frequently in the first four episodes—PPR demonstrates higher accuracy than the edge-count approach.


%% file: acl_latex.bbl
\begin{thebibliography}{19}
\providecommand{\natexlab}[1]{#1}

\bibitem[{AI(2024)}]{mistral2023largeinstruct2411}
Mistral AI. 2024.
\newblock \href {https://huggingface.co/mistralai/Mistral-Large-Instruct-2411} {Mistral-large-instruct-2411}.

\bibitem[{Alberich et~al.(2002)Alberich, Miro-Julia, and Rossello}]{alberich2002marveluniverselookslike}
R.~Alberich, J.~Miro-Julia, and F.~Rossello. 2002.
\newblock \href {https://arxiv.org/abs/cond-mat/0202174} {Marvel universe looks almost like a real social network}.
\newblock \emph{Preprint}, arXiv:cond-mat/0202174.

\bibitem[{Bamman et~al.(2013)Bamman, O{'}Connor, and Smith}]{4_personas}
David Bamman, Brendan O{'}Connor, and Noah~A. Smith. 2013.
\newblock \href {https://aclanthology.org/P13-1035/} {Learning latent personas of film characters}.
\newblock In \emph{Proceedings of the 51st Annual Meeting of the Association for Computational Linguistics}, pages 352--361, Sofia, Bulgaria. Association for Computational Linguistics.

\bibitem[{Baruah and Narayanan(2024)}]{2_attributefrommovie}
Sabyasachee Baruah and Shrikanth Narayanan. 2024.
\newblock \href {https://doi.org/10.1109/ICASSP48485.2024.10447353} {Character attribute extraction from movie scripts using llms}.
\newblock pages 8270--8275.

\bibitem[{Bearman and Stovel(2000)}]{8_nazi}
Peter~S. Bearman and Katherine Stovel. 2000.
\newblock \href {https://doi.org/10.1016/S0304-422X(99)00022-4} {Becoming a nazi: A model for narrative networks}.
\newblock \emph{Poetics}, 27(2):69--90.

\bibitem[{Chen et~al.(2017)Chen, Zhou, and Choi}]{10_chen-etal-2017-robust}
Henry~Y. Chen, Ethan Zhou, and Jinho~D. Choi. 2017.
\newblock \href {https://doi.org/10.18653/v1/K17-1023} {Robust coreference resolution and entity linking on dialogues: Character identification on {TV} show transcripts}.
\newblock In \emph{Proceedings of the 21st Conference on Computational Natural Language Learning}, pages 216--225, Vancouver, Canada. Association for Computational Linguistics.

\bibitem[{Elson et~al.(2010)Elson, Dames, and McKeown}]{11_elson-etal-2010-extracting}
David Elson, Nicholas Dames, and Kathleen McKeown. 2010.
\newblock \href {https://aclanthology.org/P10-1015/} {Extracting social networks from literary fiction}.
\newblock In \emph{Proceedings of the 48th Annual Meeting of the Association for Computational Linguistics}, pages 138--147, Uppsala, Sweden. Association for Computational Linguistics.

\bibitem[{Jeong et~al.(2025)Jeong, Kim, won Hwang, and Kim}]{jeong2025agentasjudgefactualsummarizationlong}
Yeonseok Jeong, Minsoo Kim, Seung won Hwang, and Byung-Hak Kim. 2025.
\newblock \href {https://arxiv.org/abs/2501.09993} {Agent-as-judge for factual summarization of long narratives}.
\newblock \emph{Preprint}, arXiv:2501.09993.

\bibitem[{Labatut and Bost(2019)}]{1_fictionalcharacterNet}
Vincent Labatut and Xavier Bost. 2019.
\newblock \href {https://arxiv.org/abs/1907.02704} {Extraction and analysis of fictional character networks: {A} survey}.
\newblock \emph{CoRR}, abs/1907.02704.

\bibitem[{Lee and Yeung(2012)}]{9_lee-yeung-2012-extracting}
John Lee and Chak~Yan Yeung. 2012.
\newblock \href {https://aclanthology.org/Y12-1022/} {Extracting networks of people and places from literary texts}.
\newblock In \emph{Proceedings of the 26th Pacific Asia Conference on Language, Information, and Computation}, pages 209--218, Bali, Indonesia. Faculty of Computer Science, Universitas Indonesia.

\bibitem[{Liu et~al.(2023)Liu, Mao, Luu et~al.}]{liu2023survey}
R.~Liu, R.~Mao, A.T. Luu, et~al. 2023.
\newblock \href {https://doi.org/10.1007/s10462-023-10506-3} {A brief survey on recent advances in coreference resolution}.
\newblock \emph{Artificial Intelligence Review}, 56(9):14439--14481.

\bibitem[{Newman and Girvan(2004)}]{7_communitystructure}
M.~E.~J. Newman and M.~Girvan. 2004.
\newblock \href {https://doi.org/10.1103/physreve.69.026113} {Finding and evaluating community structure in networks}.
\newblock \emph{Physical Review E}, 69(2).

\bibitem[{Niraula et~al.(2014)Niraula, Rus, Banjade, Stefanescu, Baggett, and Morgan}]{niraula-etal-2014-dare}
Nobal Niraula, Vasile Rus, Rajendra Banjade, Dan Stefanescu, William Baggett, and Brent Morgan. 2014.
\newblock \href {https://aclanthology.org/L14-1320/} {The {DARE} corpus: A resource for anaphora resolution in dialogue based intelligent tutoring systems}.
\newblock In \emph{Proceedings of the Ninth International Conference on Language Resources and Evaluation}, pages 3199--3203, Reykjavik, Iceland. European Language Resources Association.

\bibitem[{OpenAI(2023)}]{openai2023gpt4}
OpenAI. 2023.
\newblock \href {https://arxiv.org/abs/2303.08774} {Gpt-4 technical report}.
\newblock \emph{arXiv preprint arXiv:2303.08774}.

\bibitem[{OpenAI(2024)}]{openai2024gpt4o}
OpenAI. 2024.
\newblock \href {https://arxiv.org/abs/2410.21276} {{GPT-4o} system card}.
\newblock \emph{arXiv preprint arXiv:2410.21276}.

\bibitem[{Page et~al.(1999)Page, Brin, Motwani, and Winograd}]{ppr}
Lawrence Page, Sergey Brin, Rajeev Motwani, and Terry Winograd. 1999.
\newblock \href {http://ilpubs.stanford.edu:8090/422/} {The pagerank citation ranking: Bringing order to the web.}
\newblock Technical Report 1999-66, Stanford InfoLab.

\bibitem[{Touvron et~al.(2023)Touvron, Martin, Stone, Albert, Almahairi, Babaei, Bashlykov, Batra, Bhargava, Bhosale et~al.}]{touvron2023llama}
Hugo Touvron, Louis Martin, Kevin Stone, Peter Albert, Amjad Almahairi, Yasmine Babaei, Nikolay Bashlykov, Sharan Batra, Prajjwal Bhargava, Shruti Bhosale, et~al. 2023.
\newblock \href {https://arxiv.org/abs/2307.09288} {Llama 2: Open foundation and fine-tuned chat models}.
\newblock \emph{arXiv preprint arXiv:2307.09288}.

\bibitem[{Youngjoon et~al.(2024)Youngjoon, Junyoung, and Lee}]{KURE}
Jang Youngjoon, Son Junyoung, and Taemin Lee. 2024.
\newblock \href {https://github.com/nlpai-lab/KURE} {Kure-v1}.

\bibitem[{Zhao et~al.(2024)Zhao, Zhu, Xu, Li, Zhou, He, and Gui}]{3_llmfallshort}
Runcong Zhao, Qinglin Zhu, Hainiu Xu, Jiazheng Li, Yuxiang Zhou, Yulan He, and Lin Gui. 2024.
\newblock \href {https://doi.org/10.18653/v1/2024.findings-acl.454} {Large language models fall short: Understanding complex relationships in detective narratives}.
\newblock In \emph{Findings of the Association for Computational Linguistics: ACL 2024}, pages 7618--7638, Bangkok, Thailand. Association for Computational Linguistics.

\end{thebibliography}
